\documentclass[11pt]{article}
\usepackage[preprint]{acl}
\usepackage{times}
\usepackage{latexsym}
\usepackage[T1]{fontenc}
\usepackage[utf8]{inputenc}
\usepackage{CJKutf8}
\usepackage{marvosym}
\usepackage{inconsolata}

\usepackage{amsmath,amsfonts,bm}

\def\eqref#1{equation~\ref{#1}}

\def\1{\bm{1}}

\DeclareMathAlphabet{\mathsfit}{\encodingdefault}{\sfdefault}{m}{sl}
\SetMathAlphabet{\mathsfit}{bold}{\encodingdefault}{\sfdefault}{bx}{n}

\usepackage{booktabs}
\usepackage{stfloats}
\usepackage{microtype}
\usepackage{multirow}
\usepackage{graphicx}
\usepackage{subcaption}
\usepackage{xcolor}
\usepackage{tcolorbox}
\tcbuselibrary{skins,breakable}
\usepackage{fvextra}
\definecolor{matisse}{RGB}{31,94,137}
\usepackage{colortbl}
\usepackage{adjustbox}
\usepackage[inline]{enumitem}

\usepackage{fontawesome5}

\renewcommand{\arraystretch}{0.9}

\title{MemBubble: Benchmarking How Agent Memory Shapes \\ LLM Response Diversity}

\author{Jiajun Tan$^{\alpha\beta}$\hspace{0.5em} Fei Sun$^{\alpha}$\textsuperscript{\tiny{\faIcon[regular]{envelope}}} \hspace{0.2em}  {\bf Xinyu Ma}$^{\gamma}$ \hspace{0.2em} {\bf Shixuan Zhang}$^{\beta}$  \hspace{0.2em} \\ 
   \textbf{Du Su}$^{\alpha}$  \hspace{0.2em} \textbf{Qi Cao}$^{\alpha}$ \hspace{0.2em} {\bf Daiting Shi}$^{\gamma}$ \hspace{0.2em} \textbf{Huawei Shen}$^{\alpha\beta}$ \hspace{0.2em} \textbf{Xueqi Cheng}$^{\alpha\beta}$ \\
  $^{\alpha}$State Key Laboratory of AI Safety, Institute of Computing Technology, CAS\\
  $^{\beta}$University of Chinese Academy of Sciences  \hspace{1em} $^{\gamma}$ Baidu Inc.\\
 \tt{tanjiajun22s@ict.ac.cn \,\,\, \textsuperscript{\tiny{\faIcon[regular]{envelope}}}sunfei@ict.ac.cn}}

\begin{document}

\maketitle

\begingroup
\begin{NoHyper}
\renewcommand*{\thefootnote}{\tiny\faIcon[regular]{envelope}}
\footnotetext{Corresponding author: Fei Sun (\href{sunfei@ict.ac.cn}{sunfei@ict.ac.cn})}
\end{NoHyper}
\endgroup

\renewcommand*{\thefootnote}{\arabic{footnote}}

\begin{abstract}
Memory serves as the core component of LLM-based agents, enabling personalized responses via extracting and using user preferences from past interactions.
However, its use may reduce response diversity, thereby narrowing the range of information available to users.
Existing evaluations rely largely on synthetic interactions and cover few memory systems, leaving the effects of memory on response diversity insufficiently evaluated in real-world scenarios.
We introduce \textit{MemBubble}, a benchmark of 1000 instances built on preferences extracted from real user--assistant conversations.
To measure the extent to which memory concentrates response distributions, we construct test queries in scenarios distinct from user preferences while preserving topical relevance through shared preference categories.
Across multiple LLMs and memory mechanisms, entropy decreases by 34.7\% on average compared with directly providing the full interaction history, highlighting the important role of memory processing in diversity loss.
We also explore simple user-side interventions and changes in information availability but find that these do not effectively close the diversity gap.
Our findings highlight the need to consider preserving diversity and user access to information alongside improvements in personalization when designing and using agent memory systems.

\end{abstract}

\begin{figure}[!t]
\centering
\includegraphics[width=0.99\columnwidth]{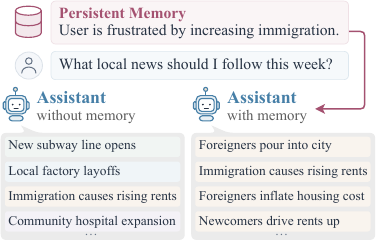}
\captionsetup{skip=6pt}
\caption{Illustration of response diversity decrease as memory steers responses toward similar preference-aligned content, potentially reinforcing existing biases.}
\label{fig:motivation}
\end{figure}

\section{Introduction}
\label{sec:introduction}

LLM-based agents increasingly rely on persistent memory to extract and utilize user preferences from past conversations~\citep{Zhang2025Survey,Hu2025Memory}.
This allows agents to provide personalized support for users with a wide range of tasks~\citep{Anthropic2025Effective}.
However, recent studies show that memory-based personalization can also cause problems such as sycophantic responses and irrelevant content~\citep{Hu2026OPBench,Xiang2026MemSycoBench}.

In this paper, we focus on a further risk: the effect of memory on response diversity~\citep{Wang2026Evaluating}.
Users increasingly turn to AI assistants for information and ideas~\citep{Chatterji2025People}.
Many of these requests are open-ended with a wide range of suitable answers.
By drawing on stored user preferences, memory can steer models toward a limited set of responses, restricting user exposure to diverse advice and perspectives~\citep{Wang2026Evaluating}.
This narrowing may in turn reinforce existing biases and fuel polarization~\citep{Sharma2024Echo}, similar to the \emph{filter bubble} effect in recommender systems~\citep{Nguyen2014Exploring,Lazovich2023Filter}.
For example, Figure~\ref{fig:motivation} illustrates a user asking for local news recommendations after expressing frustration with immigration.
Having captured this frustration in memory, the assistant may repeatedly suggest negative stories about people from abroad, potentially strengthening negative stereotypes about foreigners while overlooking other important issues.

Although prior studies have shown that user preferences can narrow the response space~\citep{Wang2026Evaluating,Feng2026Personalized}, their evaluations largely rely on synthetic interaction histories.
Such synthetic dialogues not only suffer from the homogeneity of LLM generations~\citep{Jiang2025Artificial}, but also express preferences in ways that differ from those of real users~\citep{Xiao2026AlpsBench,Guo2026Natural}.
Moreover, these studies provide preferences directly in context or through simple retrieval~\citep{Wang2026Evaluating}, leaving representative memory architectures unexamined.
These concerns motivate a systematic evaluation regarding how user preferences from natural conversations affect LLM response diversity after being processed by actual memory systems.

To this end, we introduce the \emph{MemBubble} benchmark, which aims to test whether memory narrows the range of plausible responses by constructing evaluation tasks that admit multiple valid answers while leaving room for memory to influence the response distribution.
We first filter user--assistant dialogues from widely used large-scale real conversation datasets~\citep{Zheng2023LMSYSChat1M,Zhao2024WildChat}.
LLMs then extract user preferences from the filtered dialogues and identify the preference category and corresponding scenario.
We construct open-ended queries with same category but different scenario to maintain relevance to the preferences while avoiding tasks that call for preference-aligned answers.
Each query is paired with a multi-session interaction history formed by randomly mixing the preference dialogue with unrelated background dialogues.
The final benchmark contains 1000 history--query pairs based on source dialogues in 35 languages and queries spanning 14 topic categories.

Building on this dataset, we develop an evaluation framework that approximates model output distributions by repeatedly sampling responses to each query and measures diversity over functional equivalence classes~\citep{Zhang2025NoveltyBench}.
Experiments on six LLMs and five representative memory mechanisms reveal a widespread decrease in response diversity.
Entropy over functionally distinct answers decreases by 34.7\% on average compared with directly providing the full interaction history, indicating that memory systems play an important role in diversity loss.
We also explore simple user-side interventions to response generation but find that they do not effectively close the diversity gap.
Further analysis of information availability reveals that the magnitude of diversity loss varies with interaction history length and retrieval depth, with patterns differing across memory systems.
These findings underscore the importance of preserving response diversity and user access to information in agent memory design and use as personalization capabilities advance.

Our contributions are summarized as follows:
\begin{itemize}[topsep=1pt, itemsep=0pt]
    \item We propose \textit{MemBubble}, a benchmark with 1000 instances built from real user--assistant conversations for evaluating how agent memory shapes LLM response diversity.
    \item We conduct experiments across multiple LLMs and memory mechanisms, revealing widespread diversity loss and the important role of memory systems in this phenomenon.
    \item We show that the diversity gap persists under simple adjustments to response generation and information availability, motivating attention to diversity and information access alongside personalization in memory design.
\end{itemize}

\begin{figure*}[t]
\centering
\includegraphics[width=\textwidth]{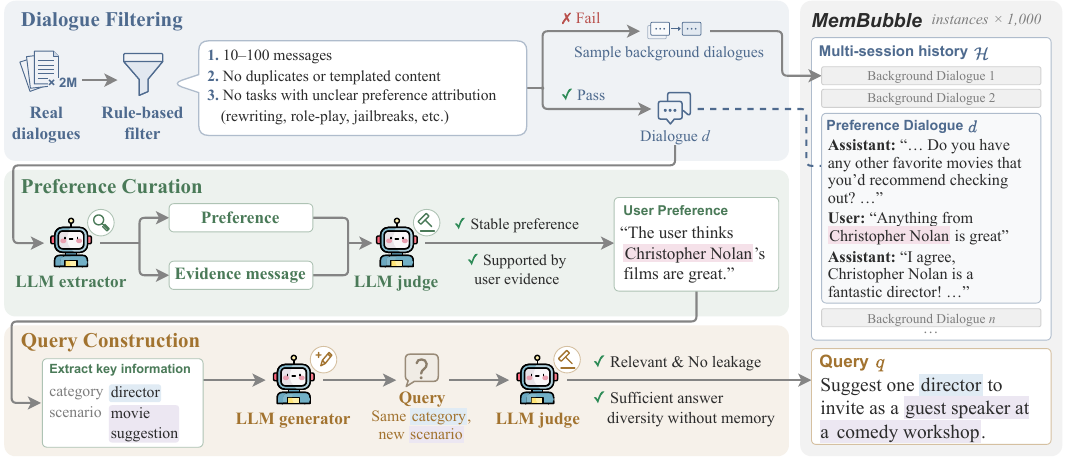}
\captionsetup{skip=4pt}
\caption{Construction of \textit{MemBubble} benchmark. Each instance pairs a multi-session interaction history assembled from a preference dialogue and sampled background dialogues with an open-ended test query.}
\label{fig:construction}
\end{figure*}

\section{Preliminaries}
\label{sec:preliminaries}

\subsection{LLM Agent Memory}
We consider an LLM agent equipped with persistent memory that captures information across multiple sessions between user and agent~\citep{Zhang2025Survey,Hu2025Memory}.
Let $\mathcal{H}=((u_i,a_i))_{i=1}^{T}$ denote the interaction history, where $u_i$ and $a_i$ are the user message and agent response at turn $i$.
For a current query $q$, memory construction $W$ and retrieval $R$ give
\begin{equation}
\mathcal{M}=W(\mathcal{H}), \qquad r=R(\mathcal{M},q).
\end{equation}
Here $W$ abstracts the memory writing and management process, and $\mathcal{M}$ stores past conversations or information extracted from them in external storage.
To answer the current query $q$, the memory system retrieves relevant memory items $r$ from $\mathcal{M}$ and adds them to the context for its LLM backbone $\pi_\theta$ to generate a response:
\begin{equation}
y\sim \pi_\theta(\cdot\mid q,r).
\end{equation}

\subsection{Response Diversity}
\label{sec:response-diversity}

For an open-ended query $q$ admitting multiple meaningful answers, we independently sample $k$ responses $Y_k=(y_1,\ldots,y_k)$ from $\pi_\theta(\cdot\mid q,r)$.
Following~\citet{Zhang2025NoveltyBench}, we define \emph{response diversity} based on distribution of responses in $Y_k$ across functional equivalence classes.
A group of responses is \emph{functionally equivalent} if, for any pair within the group, seeing one after the other provides little additional task-relevant value to the user.
This equivalence induces a partition of the sampled responses into classes, on which a diversity measure $\delta$ is defined as:
\begin{equation}
\begin{aligned}
\mathcal{Q}(Y_k;q)&:=\{C_1,\ldots,C_m\},\\
\mathrm{Diversity}(Y_k;q)&:=\delta(\mathcal{Q}(Y_k;q)).
\end{aligned}
\end{equation}

\section{Benchmark Construction}
\label{sec:data}

In this section, we describe how we construct the \textit{MemBubble} benchmark by designing evaluation tasks based on user preferences observed in real-world interactions.
To facilitate observing effects of agent memory on LLM response diversity, we argue that these tasks should admit multiple valid answers while leaving room for memory to influence the response distribution.
Those unrelated to user preferences may offer little opportunity to observe their influence, while response concentration in tasks that require preference alignment may primarily reflect appropriate personalization.
Therefore, we identify the preference category and corresponding scenario and construct test queries with same category but different scenario.
Figure~\ref{fig:construction} outlines our automated construction pipeline, with further details in Appendix~\ref{app:dataset_construction}.

\subsection{Dialogue Filtering}

\textbf{Source corpus.}
We utilize two large-scale datasets containing real user--assistant conversations: LMSYS-Chat-1M~\citep{Zheng2023LMSYSChat1M} and WildChat~\citep{Zhao2024WildChat}.
Each dataset originally contains approximately 1,000,000 samples.

\noindent \textbf{Length filtering and deduplication.}
We keep conversations containing 10--100 messages to provide sufficient context for identifying user preferences while limiting the cost of subsequent processing.
We then deduplicate conversations by hashing the concatenation of all user messages after normalizing case and whitespace.

\noindent \textbf{Template and placeholder filtering.}
We discard dialogues initiated in batches with same main prompts, differing only in a few words of variable fields and therefore missed after deduplication.
We also remove samples containing dataset-specific placeholders, as these can obscure key information about preferences.

\noindent \textbf{Task-type filtering.}
We further exclude conversations dominated by rewriting, translation, polishing, proofreading, or summarization, which typically express temporary requirements rather than stable preferences.
We also exclude role-playing and jailbreak conversations, where persona descriptions and instructions can easily be confused with the user's own preferences.
For each task category, we define exclusion rules combining keyword lists with textual features of user messages.

\begin{figure*}[t]
\centering
\begin{minipage}[b]{0.36\linewidth}
  \centering
  \footnotesize
  \setlength{\tabcolsep}{2pt}
  \renewcommand{\arraystretch}{1.12}
  \raisebox{6pt}[\height][\depth]{\begin{tabular}{@{}l@{\hspace{3pt}}r@{}}
\toprule
Statistic & Value \\
\midrule
\# Total Samples & 1000 \\
\# Samples from WildChat & 758 \\
\# Samples from LMSYS-chat-1M & 242 \\
Avg. dialogue length (messages) & 17.6 \\
Avg. dialogue length (words) & 1720.3 \\
Avg. preference length (words) & 16.2 \\
Avg. test query length (words) & 11.8 \\
\bottomrule
\end{tabular}
}
  \par\smallskip
  {\footnotesize (a) Dataset statistics}
\end{minipage}\hfill%
\begin{minipage}[b]{0.3\linewidth}
  \centering
  \parbox[c][0.83\linewidth][t]{\linewidth}{%
    \centering
    \includegraphics[width=\linewidth]{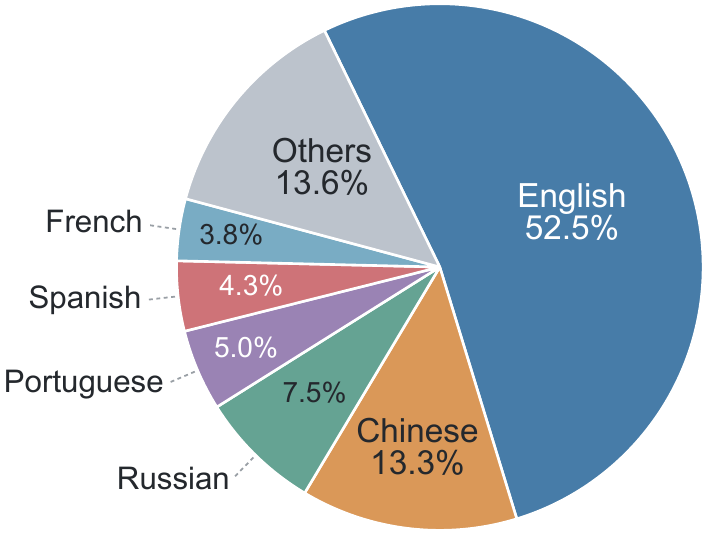}%
  }
  \par\smallskip
  {\footnotesize (b) Dialogue language distribution}
\end{minipage}\hfill%
\begin{minipage}[b]{0.3\linewidth}
  \centering
  \parbox[c][0.83\linewidth][t]{\linewidth}{%
    \centering
    \includegraphics[width=\linewidth]{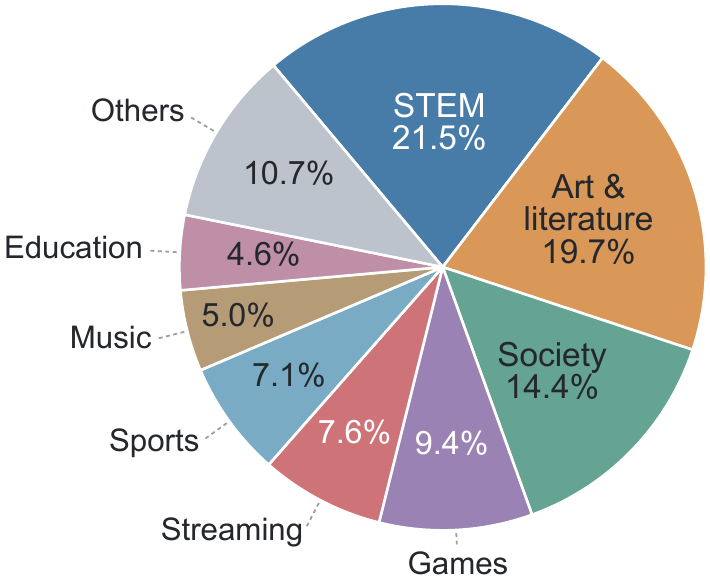}%
  }
  \par\smallskip
  {\footnotesize (c) Test task topic distribution}
\end{minipage}
\captionsetup{skip=5pt}
\caption{Overview of our proposed \textit{MemBubble} dataset.}
\label{fig:dataset-overview}
\end{figure*}

\subsection{Preference Curation}
\label{sec:pref_extract}

\textbf{Preference extraction.}
We prompt an LLM extractor with a predefined extraction template adapted from~\citet{Xiao2026AlpsBench} to extract candidate preferences.
Each memory records a preference description and supporting evidence quoted from user messages in the corresponding dialogue.
A single dialogue may yield multiple preferences, but each must be supported by explicit evidence.

\noindent \textbf{Consistency check.}
Although we require explicit evidence messages, extracted preferences may still misrepresent user statements and contain preferences beyond what the original dialogue supports.
We therefore perform an independent check using an LLM verifier from a different model family from that of the extractor~\citep{Zhao2025LanguageModelCouncil}.
The check covers two aspects: i) whether facts in preference description are consistent with the preference dialogue; and ii) whether it describes the preference with an appropriate scope, without overstating preferences that are not clearly expressed.
We discard samples that fail either of these two criteria.

\noindent \textbf{Semantic assessment.}
After verifying consistency, we use two LLM judges from different providers to independently assess the suitability of the remaining preferences for diversity evaluation.
Each judge first applies an exclusion checklist to retain clear and stable preferences compatible with our evaluation setting, excluding temporary emotional states and surface-level preferences for style and formatting.
It then checks whether the option indicated by each preference belongs to a nameable category with several plausible alternatives and whether the category can be described without revealing that option.
For preferences that pass these checks, the judge extracts the preference category $c$ and corresponding scenario $s$.
Two LLM judges from different providers assess each preference independently given the importance of this step, and we only keep samples that are accepted by both judges.

\subsection{Query Construction}
\label{sec:query_construct}

\textbf{Query generation.}
We construct test queries based on identified suitable preferences with key attribute fields. 
We prompt an LLM generator to write queries using only the category $c$ and attached scenario $s$ extracted in the previous stage, without providing the concrete preference value.
We require each query to present a single, impersonal request for one member of $c$, with a scenario that must differ substantively from $s$ without adding conditions that narrow the range of reasonable answers.

\noindent \textbf{Leakage and relevance check.} We check the relation between each generated query and its corresponding preference to ensure the query leaves room for memory to influence the response distribution.
An LLM judge verifies that the option indicated by the preference remains a  natural answer to the query and that the query does not directly name or narrowly imply that option.

\noindent \textbf{Diversity pre-test.}
To facilitate the observation of diversity changes, the query should elicit diverse responses without user preference involved.
For each candidate query, we independently sample 5 responses from an LLM.
A separate judge groups these responses according to functional equivalence as defined in \S\ref{sec:response-diversity} and computes the class entropy\footnote{See \S\ref{sec:diversity-metrics} for metric details.} over the resulting groups.
We set a entropy threshold of 1.4 bits and include samples exceeding this threshold in the final dataset.

\subsection{Building Interaction History}
\label{sec:history-construction}

In real-world applications, users may interact with agents over time, and their conversations often span different tasks and topics.
To simulate this pattern, we extend each preference dialogue in \textit{MemBubble} into a multi-session interaction history.
We draw additional background dialogues from LMSYS-Chat-1M and WildChat, excluding those used to construct the benchmark.
Then we uniformly sample 50 background dialogues in the same language as the preference dialogue and randomly remove dialogues until their combined length is no more than 30,000 tokens.
Following~\citet{Wu2024LongMemEval}, we combine the preference dialogue with these background dialogues in random order to form the final interaction history.

\subsection{Dataset Overview}

Our final dataset contains 1000 pairs of multi-session interaction histories and corresponding open-ended test queries.
Figure~\ref{fig:dataset-overview} summarizes the main characteristics of the dataset.
Preference dialogues cover 35 identified languages, with English and Chinese being the most common.
Test queries span 14 topic categories, with STEM and art \& literature forming the largest groups.
Further dataset details are provided in Appendix~\ref{app:dataset_statistics}.

\begin{table*}[t]
\newcommand{\metrichead}[1]{#1}
\newcommand{\meandelta}[2]{#1{\scriptsize\,%
\ifdim #2pt<0pt\color{green!40!black}%
\else\ifdim #2pt>0pt\color{red!65!black}\fi\fi%
\ensuremath{#2}}}
\centering
\fontsize{9pt}{11pt}\selectfont
\setlength{\tabcolsep}{1pt}
\newcommand{\memoryhead}[1]{\makebox[\linewidth][c]{#1}}
\newlength{\memorycolumnwidth}
\settowidth{\memorycolumnwidth}{MemoryBank}
\newlength{\nomemorycolumnwidth}
\settowidth{\nomemorycolumnwidth}{No mem.}
\resizebox{\textwidth}{!}{%
\begin{tabular}{@{}l@{\hspace{3.5pt}}>{\centering\arraybackslash}p{\nomemorycolumnwidth}>{\centering\arraybackslash}p{\memorycolumnwidth}!{\color{gray!55}\vrule width 0.4pt}*{5}{>{\centering\arraybackslash}p{\memorycolumnwidth}}@{}}
\toprule
& \memoryhead{No mem.} & \memoryhead{Full history} & \memoryhead{NaiveRAG} & \memoryhead{MemoryBank} & \memoryhead{A-Mem} & \memoryhead{mem0} & \memoryhead{MemoryOS} \\
\midrule
\multicolumn{8}{c}{\metrichead{Adapted  Distinct-$k$}} \\
\midrule
GPT-6-Sol & 3.67 & \meandelta{3.54}{-0.12} & \meandelta{2.67}{-1.00} & \meandelta{2.41}{-1.26} & \meandelta{2.28}{-1.39} & \meandelta{2.16}{-1.51} & \meandelta{2.81}{-0.85} \\
GPT-5.6-Luna & 3.56 & \meandelta{3.50}{-0.06} & \meandelta{2.72}{-0.85} & \meandelta{2.43}{-1.13} & \meandelta{2.23}{-1.33} & \meandelta{2.21}{-1.35} & \meandelta{2.71}{-0.85} \\
Claude Haiku 4.5 & 3.58 & \meandelta{2.87}{-0.71} & \meandelta{2.86}{-0.72} & \meandelta{2.34}{-1.24} & \meandelta{2.38}{-1.20} & \meandelta{2.09}{-1.49} & \meandelta{3.04}{-0.54} \\
Gemini 3.7 Flash & 2.30 & \meandelta{2.47}{+0.18} & \meandelta{2.13}{-0.17} & \meandelta{1.93}{-0.36} & \meandelta{1.79}{-0.51} & \meandelta{1.98}{-0.32} & \meandelta{2.11}{-0.18} \\
GLM-5.3-flash & 3.05 & \meandelta{3.55}{+0.50} & \meandelta{2.63}{-0.43} & \meandelta{2.38}{-0.68} & \meandelta{2.14}{-0.91} & \meandelta{2.19}{-0.86} & \meandelta{2.65}{-0.41} \\
Llama3.3-70B & 2.82 & \meandelta{3.15}{+0.33} & \meandelta{2.54}{-0.28} & \meandelta{1.88}{-0.94} & \meandelta{2.49}{-0.33} & \meandelta{1.67}{-1.15} & \meandelta{2.38}{-0.44} \\
\specialrule{\lightrulewidth}{\aboverulesep}{2pt}
{\fontsize{8.5pt}{11pt}\selectfont Avg. change to No mem.} & -- & +1.84\% & \textminus{}17.10\% & \textminus{}28.63\% & \textminus{}28.69\% & \textminus{}33.89\% & \textminus{}16.50\% \\
\specialrule{\lightrulewidth}{2pt}{\belowrulesep}
\multicolumn{8}{c}{\metrichead{Class Entropy}} \\
\midrule
GPT-6-Sol & 1.46 & \meandelta{1.39}{-0.06} & \meandelta{0.94}{-0.52} & \meandelta{0.79}{-0.66} & \meandelta{0.75}{-0.71} & \meandelta{0.67}{-0.78} & \meandelta{1.02}{-0.44} \\
GPT-5.6-Luna & 1.41 & \meandelta{1.35}{-0.05} & \meandelta{0.98}{-0.43} & \meandelta{0.82}{-0.58} & \meandelta{0.71}{-0.70} & \meandelta{0.71}{-0.70} & \meandelta{0.98}{-0.42} \\
Claude Haiku 4.5 & 1.37 & \meandelta{1.03}{-0.33} & \meandelta{1.04}{-0.33} & \meandelta{0.77}{-0.59} & \meandelta{0.78}{-0.58} & \meandelta{0.64}{-0.73} & \meandelta{1.12}{-0.24} \\
Gemini 3.7 Flash & 0.76 & \meandelta{0.86}{+0.09} & \meandelta{0.66}{-0.10} & \meandelta{0.56}{-0.21} & \meandelta{0.47}{-0.29} & \meandelta{0.57}{-0.19} & \meandelta{0.65}{-0.11} \\
GLM-5.3-flash & 1.14 & \meandelta{1.36}{+0.22} & \meandelta{0.91}{-0.22} & \meandelta{0.79}{-0.35} & \meandelta{0.66}{-0.47} & \meandelta{0.69}{-0.45} & \meandelta{0.93}{-0.21} \\
Llama3.3-70B & 1.00 & \meandelta{1.15}{+0.15} & \meandelta{0.87}{-0.13} & \meandelta{0.51}{-0.49} & \meandelta{0.81}{-0.19} & \meandelta{0.40}{-0.60} & \meandelta{0.81}{-0.20} \\
\specialrule{\lightrulewidth}{\aboverulesep}{2pt}
{\fontsize{8.5pt}{11pt}\selectfont Avg. change to No mem.} & -- & +2.39\% & \textminus{}22.75\% & \textminus{}39.58\% & \textminus{}39.88\% & \textminus{}46.66\% & \textminus{}21.77\% \\
\specialrule{\heavyrulewidth}{2pt}{\belowbottomsep}
\end{tabular}}
\captionsetup{skip=6pt}
\caption{Response diversity results across different memory conditions with $k{=}10$ and temperature $1.0$.
Higher values indicate greater diversity for both metrics.
Annotations to the right of the metric values report signed differences from \textit{No mem}.
}
\label{tab:main-results-independent}
\end{table*}

\section{Evaluation Settings}
\label{sec:evaluation}

Based on \textit{MemBubble} dataset, we evaluate how agent memory affects LLM response diversity.
We compare responses generated under different memory conditions using diversity metrics based on functional equivalent classes.
This section outlines the evaluation setup, and further details are provided in Appendix~\ref{app:evaluation_details}.

\subsection{Memory and Response LLMs}
\label{sec:mem-and-llms}
\textbf{Memory conditions.}
We evaluate four representative agent memory architectures: MemoryBank~\citep{Zhong2024MemoryBank}, A-Mem~\citep{Xu2025AMEM}, mem0~\citep{Chhikara2025Mem0}, and MemoryOS~\citep{Kang2025MemoryOS}.
Each architecture builds persistent memory from the multi-session history and retrieves relevant memory items to condition responses to the test query.
We also include NaiveRAG~\citep{Xiang2026MemSycoBench} as a simple memory baseline that retrieves relevant sessions from the original interaction history.
We use two reference conditions: the input context contains only the query under \textit{No mem.} and additionally the complete interaction history under \textit{Full history}.

\noindent \textbf{Response Models.}
We consider four closed-source LLMs from different providers and  different tiers: GPT-6-Sol~\citep{OpenAI2026GPT6Sol}, GPT-5.6-Luna~\citep{OpenAI2026GPT56}, Claude Haiku 4.5~\citep{Anthropic2025ClaudeHaiku45}, and Gemini 3.7 Flash~\citep{Google2026Gemini37Flash}.
We also include two open-source LLMs: GLM-5.3-flash~\citep{ZAI2026GLM53Flash} and Llama3.3-70B~\citep{Meta2024Llama33}.
We report results for all six models in the main experiment and use GPT-5.6-Luna for other experiments unless otherwise specified.
For each query and memory condition, we independently sample $k=10$ responses at temperature 1.0 with a maximum output length of 512 tokens.

\begin{figure*}[t]
\centering
\includegraphics[width=\textwidth]{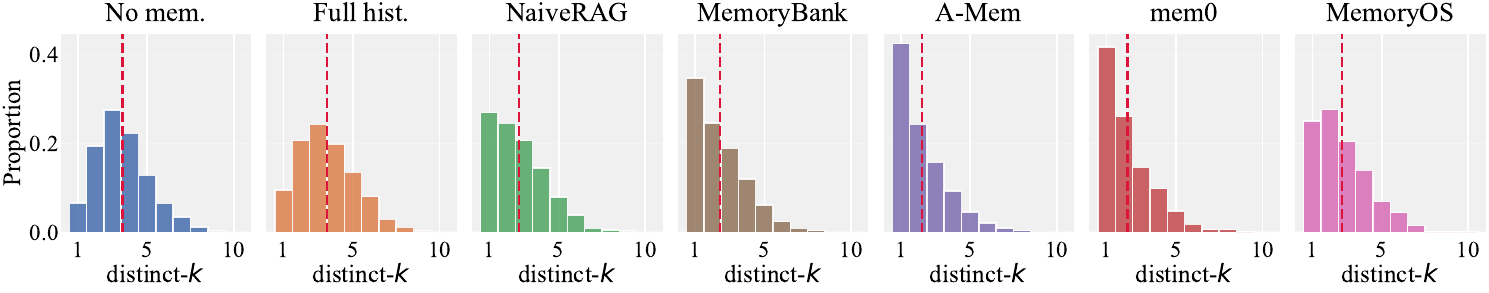}
\captionsetup{skip=6pt}
\caption{Distribution of distinct-$k$ of \textit{MemBubble} samples across memory conditions. Dashed vertical lines indicate the mean for each condition.}
\label{fig:distinct-k-distribution}
\end{figure*}

\subsection{Diversity Evaluation}
\label{sec:diversity-metrics}

To quantify response diversity, we use adapted $\text{distinct-}k$ and class entropy based on definition in Section~\ref{sec:response-diversity} as metrics.

\noindent \textbf{Adapted distinct-$k$.}
Distinct-$k$ $d_k$ is defined as the number of functional equivalence classes among $k$ responses independently generated for the same query $q$~\citep{Zhang2025NoveltyBench}.
To improve classification accuracy and reduce the overhead of pairwise comparisons, we use an LLM judge to directly partition all $k$ responses into functional equivalence classes instead of binary classifier in original implementation in~\citet{Zhang2025NoveltyBench}.
Given the resulting partition $\mathcal{Q}(Y_k;q)=\{C_1,\ldots,C_m\}$, the metric is calculated by
\begin{equation}
d_k:=|\mathcal{Q}(Y_k;q)|=m.
\end{equation}
Higher $d_k$ indicates broader coverage of functionally different answers.

\noindent \textbf{Class entropy.}
This metric complements distinct-$k$ by capturing differences in response concentration that the class count alone does not reflect.
Let $p_j=|C_j|/k$ denote the proportion of responses assigned to class $C_j$.
We define class entropy $H$ from these proportions as
\begin{equation}
H:=-\sum_{j=1}^{m} p_j\log_2 p_j.
\end{equation}
Higher $H$ indicates a less concentrated response distribution, i.e. higher response diversity.

\subsection{Implementation Details}
\label{sec:experimental-setup}

We use DeepSeek-V4-Flash~\citep{DeepSeek2026V4} as the shared writer for persistent memory construction, with details provided in Appendix~\ref{app:baseline_implementations}.
Once built, memory will be reused across response generation models for fair comparison.
We use text-embedding-3-small~\citep{OpenAITextEmbedding3Small} for retrieval across all baselines, with top-$k$ and other hyperparameters following the official settings of each baseline, detailed in Appendix~\ref{app:retrieval_generation_settings}.
DeepSeek-V4-Flash also serves as the judge for partitioning responses into functional equivalence classes, with prompt provided in Appendix~\ref{app:metric_computation}.

\section{Experimental Results and Analysis}
\label{sec:results}

In this section, we apply the evaluation protocol from \S\ref{sec:evaluation} on our proposed \textit{MemBubble} benchmark to empirically examine how agent memory shapes LLM response diversity, guided by the following three questions:
\textbf{Q1:} To what extent does memory affect response diversity? (\S\ref{sec:diversity-prevalence})
\textbf{Q2:} Can simple user interventions mitigate diversity reduction? (\S\ref{sec:user-mitigation})
\textbf{Q3:} How does information availability affect response diversity? (\S\ref{sec:history-retrieval-effects})
Further analyses of these results are provided in Appendix~\ref{app:supplementary_results}.

\subsection{Q1: To what extent does memory affect response diversity?}
\label{sec:diversity-prevalence}

We first assess how memory affects LLM response diversity by comparing five memory mechanisms against  references.
Table~\ref{tab:main-results-independent} shows average reductions of 25.0\% in distinct-$k$ and 34.1\% in class entropy across the six response models and five memory mechanisms compared to \textit{No mem.}, and distinct-$k$ decreases by 25.7\% and class entropy by 34.7\% compared to \textit{Full history}.
Figure~\ref{fig:distinct-k-distribution} shows a corresponding shift toward fewer functional equivalence classes: about 34\% of samples have $d_k=1$ on average across memory mechanisms, compared with about 6\% under \textit{No mem.} and 9.5\% under \textit{Full history}.
These results show widespread reductions in response diversity and suggest that memory processing plays an important role in this reduction.

\begin{figure}[t]
\centering
\includegraphics[width=\linewidth]{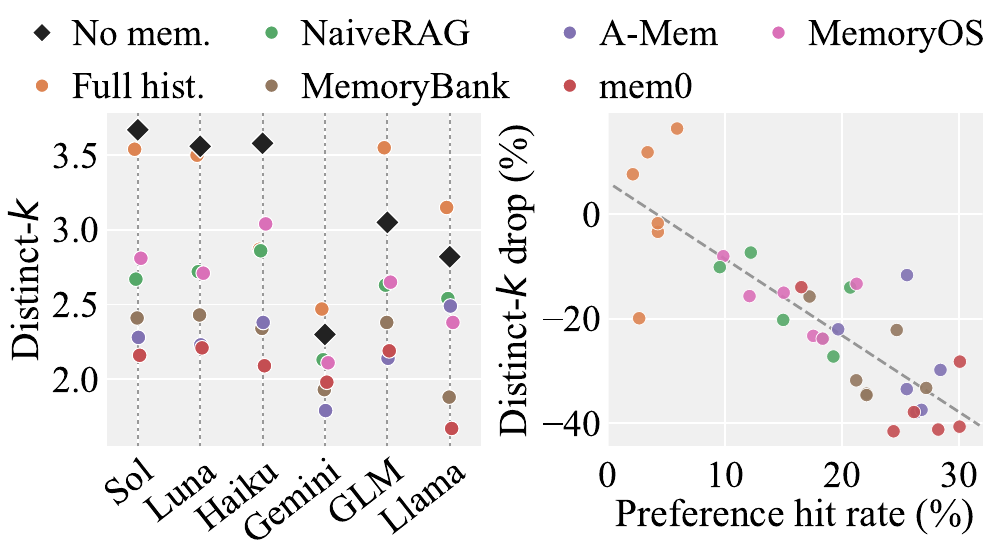}
\captionsetup{skip=6pt}
\caption{Left: mean distinct-$k$ across models and memory conditions. Right: distinct-$k$ change from \textit{No mem.} versus preference hit rate. One point represents a model with a memory condition.}
\label{fig:distinct-k-drop}
\end{figure}

The magnitude of this reduction varies across memory mechanisms and response models.
mem0 produces the largest average reductions in both metrics, while MemoryOS has the mildest effect.
MemoryBank and A-Mem also concentrate the distribution at low distinct-$k$ values, whereas MemoryOS retains a broader spread.
Among response models, GPT-6-Sol shows the largest average reductions in both metrics, while Gemini 3.7 Flash shows the smallest.
Figure~\ref{fig:distinct-k-drop} (left) shows that memory mechanisms bring distinct-$k$ to similarly low levels across response models despite differences in diversity under \textit{No mem.}
In the right panel, larger diversity reductions coincide with higher preference hit rates, indicating that reduced diversity accompanies more frequent preference-aligned responses even when the query does not require such alignment.

\subsection{Q2: Can Simple User Interventions Mitigate Diversity Reduction?}
\label{sec:user-mitigation}

We next examine whether users can recover response diversity through simple changes to prompting and generation settings.
Due to API limitations on parameter adjustment, we use GLM-5.3-flash for the  temperature experiment and GPT-6-Sol for the reasoning-effort experiment.

\begin{figure}[t]
\centering
\includegraphics[width=\linewidth]{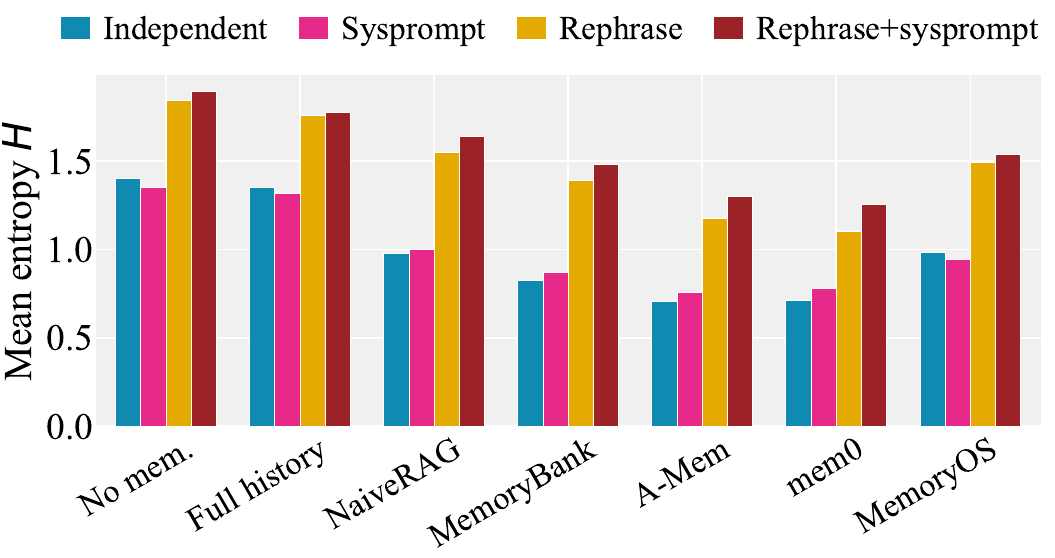}
\captionsetup{skip=6pt}
\caption{Effect of prompting strategy on mean class entropy of different memory conditions.}
\label{fig:prompting-comparison}
\end{figure}

\noindent \textbf{Prompting strategy.}
We adopt two prompting strategies from \citet{Zhang2025NoveltyBench}.
For \textit{Rephrase}, an LLM rewrites the query into ten semantically equivalent variants, and we generate one response to each variant.
For \textit{Sysprompt}, we add an instruction to the system prompt requesting diverse responses.
We also evaluate the combination of these two strategies and compare these with the default setting of \textit{independently} sampling responses.
Figure~\ref{fig:prompting-comparison} shows that \textit{Rephrase} increases class entropy across all conditions, although responses pooled across queries cannot be directly compared with repeated samples from one query.
\textit{Sysprompt} alone has little effect.
Combining the two yields a small further increase over \textit{Rephrase}, but all five memory mechanisms remain below both \textit{No mem.} and \textit{Full history} under the same prompting strategy.

\begin{figure}[t]
\centering
\captionsetup[subfigure]{font=footnotesize}
\begin{subfigure}{\linewidth}
  \centering
  \includegraphics[width=\linewidth]{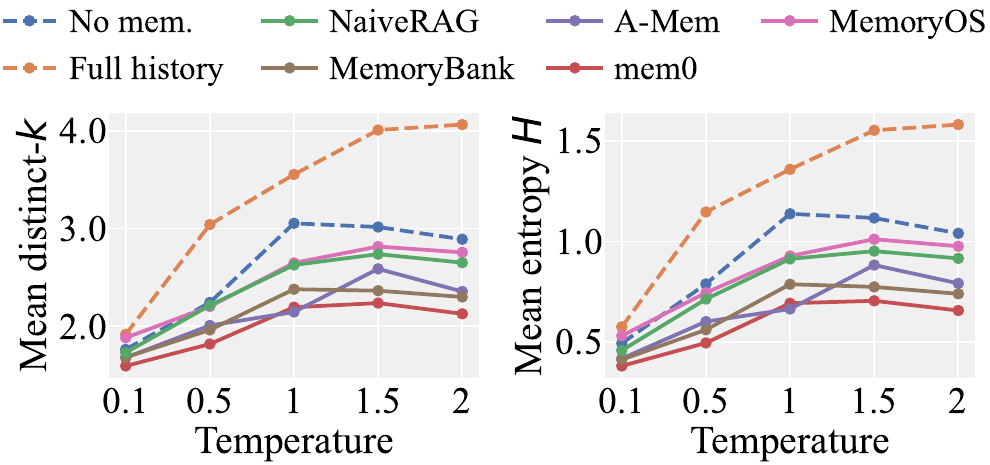}
  \caption{Sampling temperature}
  \label{fig:temperature}
\end{subfigure}
\par\medskip
\begin{subfigure}{\linewidth}
  \centering
  \includegraphics[width=\linewidth]{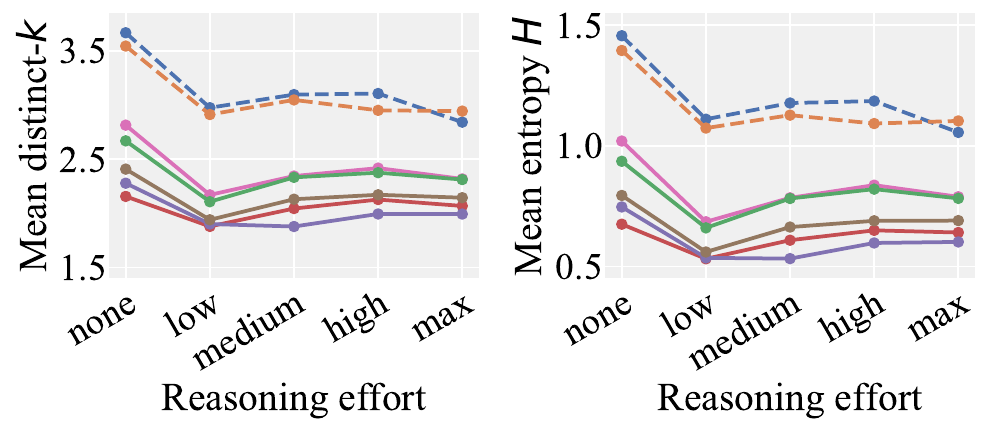}
  \caption{Reasoning effort}
  \label{fig:reasoning-effort}
\end{subfigure}
\captionsetup{skip=6pt}
\caption{Effects of generation settings on mean distinct-$k$ (left) and class entropy (right) across memory conditions: (a) sampling temperature for GLM-5.3-flash; (b) reasoning effort for GPT-6-Sol.}
\label{fig:generation-settings}
\end{figure}

\noindent \textbf{Sampling temperature.}
Higher sampling temperatures flatten the token probability distribution and are therefore expected to increase response diversity.
We vary the sampling temperature from 0.1 to 2.0 to test whether this can mitigate memory-induced diversity reduction.
Figure~\ref{fig:generation-settings}(\subref{fig:temperature}) shows similarly low diversity at low sampling temperatures, followed by an increase as temperature rises.
For all five memory mechanisms, increasing temperature beyond 1.0 yields limited diversity gains and does not close their gap to the \textit{No mem.} and \textit{Full history} baselines.

\noindent \textbf{Reasoning effort.}
We adjust the reasoning level from none to max to assess whether stronger reasoning will affect response diversity.
Figure~\ref{fig:generation-settings}(\subref{fig:reasoning-effort}) shows that  diversity tends to decrease as reasoning effort increases.
However, the magnitude of diversity reduction is overall limited, and the relative trend of memory conditions remains unchanged.

\subsection{Q3: How Does Information Availability Affect Response Diversity?}
\label{sec:history-retrieval-effects}

The range of information available to memory systems can vary at different stages.
For example, interaction histories differ in length across users, while agents may retrieve more or fewer relevant memories when generating responses.
We examine how these variations affect response diversity.

\noindent \textbf{Interaction history.}
To test how interaction history affects response diversity, we vary the initial number of background dialogues over $\{0,5,10,20,50,100\}$ while retaining the preference dialogue.
Due to cost considerations, we conduct the experiment on a randomly selected subset of 200 samples.
Table~\ref{tab:fablate-exp2-luna} shows that diversity reductions generally shrink as background dialogues increase, with occasional reversals for mem0 and MemoryOS.
Diversity remains below \textit{No mem.} throughout, suggesting that broader interaction histories can moderate memory-induced diversity reduction but do not eliminate it.

\noindent \textbf{Memory retrieval.}
To assess how retrieval depth affects response diversity, we vary retrieval top-$k$ over $\{1,3,5,10,20\}$ for representative memory systems.
Table~\ref{tab:varytopk-distinctk} shows that response diversity increases overall as more memories are retrieved, although this trend is non-monotonic and varies across systems.
However, this increase only modestly narrows the gap with \textit{No mem.}, which remains evident across all tested retrieval depths.

\begin{table}[tb]
\centering
\fontsize{9}{11}\selectfont
\setlength{\tabcolsep}{2pt}
\begin{tabular*}{\columnwidth}{@{\extracolsep{\fill}}lcccccc@{}}
\toprule
\# Dialogues & 0 & 5 & 10 & 20 & 50 & 100 \\
\midrule
MemoryBank & -1.31 & -1.00 & -0.99 & -0.90 & -0.83 & -0.67 \\
A-Mem & -1.39 & -1.24 & -1.15 & -0.98 & -0.93 & -0.54 \\
mem0 & -1.24 & -1.21 & -1.18 & -1.27 & -1.15 & -0.93 \\
MemoryOS & -1.27 & -0.82 & -0.88 & -0.80 & -0.49 & -0.40 \\
\bottomrule
\end{tabular*}
\captionsetup{skip=6pt}
\caption{Mean Distinct-$k$ change from No mem. across different background dialogues on subset from 200 randomly selected samples.}
\label{tab:fablate-exp2-luna}
\end{table}

\begin{table}[tb]
\centering
\fontsize{9}{11}\selectfont
\setlength{\tabcolsep}{2pt}
\begin{adjustbox}{max width=\textwidth}
\begin{tabular*}{\columnwidth}{@{\extracolsep{\fill}}lccccc@{}}
\toprule
top-$k$ & 1 & 3 & 5 & 10 & 20 \\
\midrule
MemoryBank & -1.15 & -1.13 & -1.13 & -1.06 & -1.00 \\
A-Mem & -1.29 & -1.31 & -1.33 & -1.37 & -1.07 \\
mem0 & -1.13 & -1.23 & -1.24 & -1.35 & -1.16 \\
NaiveRAG & -1.27 & -1.16 & -0.85 & -0.92 & -0.88 \\
\bottomrule
\end{tabular*}
\end{adjustbox}
\captionsetup{skip=6pt}
\caption{Mean distinct-$k$ change from \textit{No mem.} across retrieval top-$k$ settings.}
\label{tab:varytopk-distinctk}
\end{table}

\section{Related Work}
\label{sec:related_work}

\textbf{Agent memory.}
LLM agents use memory to keep user preferences and personalize responses~\citep{Zhang2025Survey,Xu2026Personalized}.
Representative systems summarize useful information from past user interactions~\citep{Zhong2024MemoryBank,Chhikara2025Mem0} and store it externally in forms such as user profiles~\citep{Westhaeusser2025Enabling} or linked memory notes~\citep{Xu2025AMEM} for later retrieval.

\noindent \textbf{LLM response diversity.}
Existing research has revealed limited diversity in LLM-generated responses~\citep{Padmakumar2024Writing,Sourati2026Shrinking}, showing substantial homogeneity in generations both within a single model and across different models~\citep{Zhang2025NoveltyBench,Jiang2025Artificial}.
Human studies further report reduced collective diversity in AI-assisted writing~\citep{Doshi2024Generative,Meincke2025ChatGPT} and increased selective exposure in conversational search, especially when models reinforce user views~\citep{Sharma2024Echo}.
Most of these works focus on direct model generation and do not investigate how agent memory affects response diversity.

\noindent \textbf{Over-personalization of memory-augmented LLMs.}
Memory substantially enhances personalization in LLM assistants~\citep{Zhang2026PersonaAgent,Westhaeusser2025Enabling}.
However, research has identified over-personalization as a potential risk~\citep{Lazovich2023Filter,Hu2026OPBench}.
Most studies pay attention to sycophancy~\citep{Cheng2026Sycophantic}, where models prioritize agreement with users over truthfulness~\citep{Sharma2024Sycophancy,Xiang2026MemSycoBench}, and irrelevant memory usage in generated content~\citep{Hu2026OPBench,Feng2026Personalized}.
Regarding memory-induced response diversity loss, existing evaluations largely rely on synthetic interaction histories~\citep{Wang2026Evaluating,Huang2026Controllable} and provide preferences directly in context or through simple retrieval~\citep{Wang2026Evaluating}, limiting their coverage of risks in real-world applications.

\section{Discussion and Conclusion}
\label{sec:conclusion}

Our study focuses on measuring the effects of memory on LLM response diversity, specifically the extent to which user preferences from real conversations influence model output distributions after being processed by memory systems.
To observe this influence, our dataset construction requires the scenarios of test queries to be distinct from those associated with user preferences, aiming at avoiding tasks that clearly call for memory usage.
Nevertheless, the applicability of preferences stored in memory to  specific samples may still be open to debate.
We argue that this should be comprehensively assessed by considering both the task and the intent of the user, which constitutes a separate research question beyond the scope of this work.

In this paper, we introduced the \textit{MemBubble} benchmark to systematically evaluate how user preferences from natural conversations affect LLM response diversity after being processed by actual memory systems.
We find that memory mechanisms lead to widespread reductions in response diversity across models and tasks.
We also try simple user-side interventions and adjustments to information availability but find that these do not effectively close the diversity gap.
Our findings highlight the need to weigh improvements in personalization against decrease in response diversity in agent memory systems to preserve user access to diverse information and perspectives.

\section*{Limitations}

This work primarily examines memory-induced diversity loss on open-ended tasks.
We leave a detailed investigation of the mechanisms underlying this phenomenon and the development of effective mitigation methods for future work.
Additionally, our evaluation uses fixed interaction histories and does not capture the dynamic evolution of response diversity under continued user interaction and memory updates.

\section*{Ethical Considerations}
The real user--assistant conversations used in our experiments were drawn from publicly released datasets and do not raise any privacy concerns.
We used generative AI to polish the manuscript and assist in preparing figures and tables.
As described above, LLMs also participated in dataset construction and response evaluation processes.
We do not use AI to develop the core contributions of this work, including the identification of research problem, the design of dataset construction and evaluation pipelines.

\bibliography{references}

\clearpage
\appendix
\setcounter{topnumber}{1}
\setcounter{dbltopnumber}{2}
\renewcommand{\topfraction}{0.3}
\renewcommand{\dbltopfraction}{0.42}
\setlength{\textfloatsep}{7pt plus 2pt minus 2pt}
\setlength{\dbltextfloatsep}{7pt plus 2pt minus 2pt}
\setlength{\floatsep}{7pt plus 2pt minus 2pt}
\setlength{\dblfloatsep}{7pt plus 2pt minus 2pt}
\captionsetup{skip=6pt}
\setcounter{bottomnumber}{0}
\setcounter{dblbotnumber}{0}
\makeatletter
\def\fps@figure{t}
\def\fps@table{t}
\setlength{\@fptop}{0pt}
\setlength{\@dblfptop}{0pt}
\global\@dbltopnum=0
\makeatother
\suppressfloats[t]
\section{Dataset Construction Details}
\label{app:dataset_construction}

\subsection{Rule-based Filtering Details}
\label{app:rule_based_filtering}

We apply the same filtering pipeline to LMSYS-Chat-1M and WildChat.
For template filtering, we group messages from the user by the MD5 hash of the first 200 characters of the opening user message, considering only openings of at least 60 characters.
Within each group, we retain the conversation with the most messages, breaking ties by the lexicographically smallest conversation ID.

\textbf{Placeholder filtering.}
We remove samples containing dataset-specific anonymization markers, as these can obscure key information about preferences.
For LMSYS-Chat-1M, we match \texttt{NAME\_\textbackslash d+}, such as \texttt{NAME\_1}.
For WildChat, we match \texttt{<PRESIDIO\_ANONYMIZED\_[A-Z\_]+>}.

\textbf{Task-type filtering.}
We apply case-insensitive matching over user messages using the keywords and textual features listed in Table~\ref{tab:rule_based_filters}.
For rewriting, translation, polishing, proofreading, or summarization, a user message must contain a trigger within its first 200 characters and a long text block satisfying one of the listed length conditions.
We exclude a conversation if its first user message meets these criteria or at least 50\% of its user messages do.
For role-playing and jailbreak filtering, any user message matching a listed expression or containing at least two asterisk-delimited spans of 2--60 characters triggers exclusion.
We also exclude conversations whose first user message contains at least 800 characters and at least three persona-field lines.

\begin{table*}[t]
\centering
\small
\renewcommand{\arraystretch}{1.15}
\begin{CJK*}{UTF8}{gbsn}
\begin{tabular}{@{}p{0.20\textwidth}p{0.77\textwidth}@{}}
\toprule
\textbf{Task type} & \textbf{Keywords, patterns, or conditions} \\
\midrule
Rewriting, translation, polishing, proofreading, or summarization &
\textbf{Trigger keywords:}
\texttt{translate into}, \texttt{translate}, \texttt{rewrite}, \texttt{reword}, \texttt{rephrase}, \texttt{paraphrase}, \texttt{summarize}, \texttt{summarise}, \texttt{proofread}, \texttt{correct the grammar}, \texttt{fix the grammar}, \texttt{check the grammar}, \texttt{make it more}, \texttt{make it less}, \texttt{shorten}, \texttt{expand this}; 改写, 翻译成, 翻译, 润色, 校对, 总结, 概括, 精简.
\par\smallskip
\textbf{Required long text block:}
At least 400 characters in the message; three consecutive nonempty lines totaling at least 300 characters; or a quoted span of at least 200 characters enclosed by \texttt{"""}\ldots\texttt{"""}, \texttt{"}\ldots\texttt{"}, or 「\ldots」. Any one condition suffices. \\
\midrule
Role-playing and jailbreak &
\textbf{Expressions:}
\texttt{DAN}, \texttt{AIM}, \texttt{JailBreak}, \texttt{developer mode}, \texttt{do anything now};
\texttt{you are [now] (a|an|my)}, \texttt{pretend to be}, \texttt{pretend you are}, \texttt{act as (a|an|my)};
\texttt{role-play}, \texttt{role\_play}, \texttt{roleplay};
\texttt{stay in character}, \texttt{ignore [all] (previous|prior) instructions}, \texttt{without any restrictions};
你现在是, 扮演, 不受[任何]限制.
\par\smallskip
\textbf{Persona-field prefixes:}
\texttt{name:}, \texttt{age:}, \texttt{gender:}, \texttt{personality:}, \texttt{appearance:}, \texttt{background:}, \texttt{species:}. A persona-field line begins with one of these prefixes. \\
\bottomrule
\end{tabular}
\end{CJK*}
\caption{Keywords and textual conditions used for task-type filtering, grouped by the task types excluded in Section~\ref{sec:data}. For readability, we list only English and Chinese keywords here, while the filtering rules also include keywords in other major languages. Square brackets indicate optional text, and a vertical bar separates alternatives. The conversation-level exclusion criteria are given in Appendix~\ref{app:rule_based_filtering}.}
\label{tab:rule_based_filters}
\end{table*}

\subsection{Semantic Assessment Details}
\label{app:semantic_assessment}

Each judge reviews the extracted preference and its supporting evidence, together with metadata such as the time scope and conversation language.
Personal background facts and transient states are excluded, as are isolated requests that do not establish a stable preference.
We also exclude preferences limited to response form or transforming supplied text, along with language or cultural defaults inferred from the conversation.
Unsafe material and tasks requiring the model to output code or images are ineligible.

For a preference to pass, its indicated option must belong to a natural category $c$ with several plausible alternatives.
The category name must contain at most ten words and must not repeat words from the indicated option.
For example, a preference for writing AWS Lambda functions in Python yields \emph{programming language} as $c$ and \emph{AWS Lambda function} as the original scenario $s$, with Java and Go as alternative members of the category.
Entries for which no suitable category or open-ended task can be formed are rejected.
Query construction uses only $c$ and $s$, without access to the indicated option.

GPT-5.5 and Claude Opus 4.8 independently assess each preference.
Only preferences accepted by both judges proceed to query construction.
For these preferences, we use the category $c$ and scenario $s$ extracted by GPT-5.5.
Figure~\ref{fig:semantic_assessment_prompt} provides the complete checklist and prompt.

\subsection{Models and Prompts}
\label{app:dataset_construction_prompts}

Table~\ref{tab:construction_models} lists the model and prompt used for each LLM role in the construction pipeline.
In general, each stage uses a fixed model choice for all samples.
Stages that are central to the dataset design use top-tier commercial models, while auxiliary stages use lower-cost or open-weight models.

\begin{table*}[t]
\centering
\small
\renewcommand{\arraystretch}{1.15}
\begin{tabular}{llll}
\toprule
\textbf{Stage} & \textbf{Role} & \textbf{Model} & \textbf{Prompt} \\
\midrule
\S\ref{sec:pref_extract} Preference extraction & Extractor & DeepSeek-V4-Flash~\citep{DeepSeek2026V4} & Figure~\ref{fig:memory_extraction_prompt} \\
\cmidrule(lr){1-4}
\S\ref{sec:pref_extract} Consistency check & Verifier & GPT-5.6-Luna~\citep{OpenAI2026GPT56} & Figure~\ref{fig:consistency_check_prompt} \\
\cmidrule(lr){1-4}
\multirow{2}{*}{\S\ref{sec:pref_extract} Semantic assessment} & \multirow{2}{*}{Judges} & GPT-5.5~\citep{OpenAI2026GPT55} & \multirow{2}{*}{Figure~\ref{fig:semantic_assessment_prompt}} \\
 & & Claude Opus 4.8~\citep{Anthropic2026ClaudeOpus48} & \\
\cmidrule(lr){1-4}
\S\ref{sec:query_construct} Test query generation & Generator & GPT-5.5~\citep{OpenAI2026GPT55} & Figure~\ref{fig:query_generation_prompt} \\
\cmidrule(lr){1-4}
\S\ref{sec:query_construct} Leakage and relevance check & Judge & GPT-5.5~\citep{OpenAI2026GPT55} & Figure~\ref{fig:leakage_answerability_prompt} \\
\cmidrule(lr){1-4}
\multirow{2}{*}{\S\ref{sec:query_construct} Diversity pre-test} & Response LLM & GPT-5.6-Luna~\citep{OpenAI2026GPT56} & -- \\
 & Judge & DeepSeek-V4-Flash~\citep{DeepSeek2026V4} & Figure~\ref{fig:diversity_evaluation_prompt} \\
\bottomrule
\end{tabular}
\caption{LLMs and prompts used at each stage of \textit{MemBubble} construction.}
\label{tab:construction_models}
\end{table*}

\subsection{Quality Control and Human Validation}
\label{app:dataset_quality_control}
We randomly sample 200 candidates from each selection step listed in Table~\ref{tab:construction_human_validation}, including accepted and rejected items.
Two experienced annotators independently assess whether each candidate meets the corresponding criteria, without seeing the LLM judgments.
We report pairwise agreement between the annotators and between each annotator and the LLM.
Semantic assessment uses the joint decision of the two LLM judges.

\begin{table}[t]
\centering
\small
\setlength{\tabcolsep}{3pt}
\begin{tabular}{lccc}
\toprule
\textbf{Stage} & \textbf{H1--H2} & \textbf{LLM--H1} & \textbf{LLM--H2} \\
\midrule
Consistency & 93.5\% & 86.0\% & 90.5\% \\
Semantic assessment & 81.0\% & 83.5\% & 78.5\% \\
Leakage and Relevance & 90.5\% & 87.5\% & 91.0\% \\
\bottomrule
\end{tabular}
\caption{Human validation of dataset construction. Each cell reports agreement rate, and H1 and H2 denote the two independent annotators.}
\label{tab:construction_human_validation}
\end{table}

\subsection{Interaction History Construction}
\label{app:history_memory}

The preference dialogue serves as the \emph{evidence session} in each history.
We build background pools from LMSYS-Chat-1M and WildChat, excluding dialogues used to construct \textit{MemBubble}.
Candidates flagged by the source moderation annotations or containing template spam are removed, as are those with more than three redaction placeholders.
We deduplicate candidates by message content and retain dialogues with 2--100 messages and 50--20,000 tokens.
Reservoir sampling limits each language-specific pool to 800 dialogues.

For each instance, we uniformly sample 50 background dialogues from the corresponding language pool, then randomly remove whole dialogues until their combined length is at most 30,000 tokens.
We use the \texttt{o200k\_base} encoding in \texttt{tiktoken}~\citep{OpenAITiktoken} for all background-dialogue token counts.
The budget applies only to the background dialogues; the preference dialogue is retained in full without truncation.
We randomly shuffle all sessions and record the position of the preference dialogue in the resulting history.

\subsection{Detailed Dataset Statistics}
\label{app:dataset_statistics}
\begin{figure*}[t]
\centering
\includegraphics[width=\linewidth]{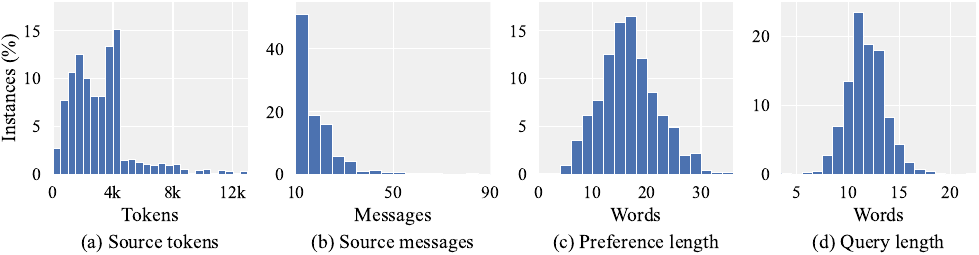}
\captionsetup{skip=5pt}
\caption{Length distributions across the 1,000 \textit{MemBubble} instances. Each panel includes all instances; bar heights indicate the percentage in each bin. Bin widths are 500 tokens, 5 messages, 2 words, and 1 word, respectively. Source dialogue lengths are measured before adding background sessions.}
\label{fig:dataset-length-distributions}
\end{figure*}

\begin{table*}[t]
\centering
\small
\setlength{\tabcolsep}{7pt}
\renewcommand{\arraystretch}{1.12}
\begin{tabular}{lrrrrr}
\toprule
\textbf{Field} & \textbf{Mean} & \textbf{Median} & \textbf{Q1--Q3} & \textbf{95th pct.} & \textbf{Min--Max} \\
\midrule
Source dialogue (tokens) & 3,118.2 & 2,914 & 1,633--4,009 & 7,236 & 76--12,675 \\
Source dialogue (messages) & 17.6 & 14 & 12--20 & 34 & 10--84 \\
Preference (words) & 16.2 & 16 & 13--19 & 25 & 1--35 \\
Test query (words) & 11.8 & 12 & 11--13 & 15 & 4--22 \\
\bottomrule
\end{tabular}

\caption{Detailed length statistics over all 1,000 instances. Q1 and Q3 are the 25th and 75th percentiles; percentiles are rounded to integers. A message is one user or assistant utterance. Preference and query word counts use whitespace splitting, consistent with the dataset overview.}
\label{tab:dataset-detailed-statistics}
\end{table*}

We report instance-weighted statistics for the 1,000 retained instances, drawn from 985 distinct source dialogues.
Source dialogue lengths refer to the session before background-session augmentation (Appendix~\ref{app:history_memory}).
We use the recorded source token counts and count each user or assistant utterance as one message.

\textbf{Length distributions.}
Figure~\ref{fig:dataset-length-distributions} shows the full distributions, with summary statistics in Table~\ref{tab:dataset-detailed-statistics}.
Source dialogues have a median length of 2,914 tokens and an interquartile range of 1,633--4,009 tokens.
The distribution has a longer upper tail: 89.6\% of instances contain at most 5,000 source tokens, and 98.3\% contain at most 10,000.
The two corpora contribute different length profiles, with median source lengths of 3,485.5 tokens for WildChat and 1,574 for LMSYS-Chat-1M.
For message counts, the 95th percentile is 34 messages.
The preference descriptions have an interquartile range of 13--19 words, while test queries are more concentrated at 11--13 words.

\begin{figure}[t]
\centering
\includegraphics[width=\linewidth]{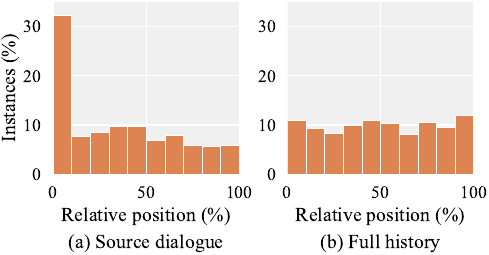}
\captionsetup{skip=5pt}
\caption{Distributions of relative evidence positions: (a) the supporting message within its source dialogue; (b) the evidence session within the constructed history. Both histograms use 10\%-wide bins over all 1,000 instances and share the same axis scales.}
\label{fig:dataset-evidence-distribution}
\end{figure}

\textbf{Evidence positions.}
Figure~\ref{fig:dataset-evidence-distribution} shows the distributions within source dialogues and within the constructed histories.
We normalize each zero-based index by the corresponding sequence length minus one: $r_m=i/(L-1)$ for supporting message $i$ in a source dialogue of $L$ messages, and $r_s=j/(S-1)$ for evidence session $j$ in a history of $S$ sessions.
Preference evidence appears in the first message for 30.4\% of instances and in the first fifth of the source dialogue for 40.0\%.
The constructed histories contain an average of 37.7 sessions.
Evidence sessions are spread throughout the history, with each 10\% interval containing 8.2--11.9\% of instances.

\section{Evaluation Details}
\label{app:evaluation_details}

\subsection{Baseline Description}
\label{app:baseline_implementations}

\noindent\textbf{MemoryBank}~\citep{Zhong2024MemoryBank} maintains conversation records alongside hierarchical event summaries and user portraits inferred from past interactions.
It encodes dialogue turns and event summaries for dense retrieval of relevant memories.
An optional updating mechanism follows the Ebbinghaus forgetting curve, allowing memory strength to decay over time and reinforcing memories when they are recalled.

\smallskip
\noindent\textbf{A-Mem}~\citep{Xu2025AMEM} organizes interactions into interconnected memory notes inspired by the Zettelkasten method.
Each note retains the original content with generated semantic annotations, including a contextual description.
An LLM links new notes to related memories and updates existing notes as additional context becomes available.
At retrieval time, semantic matching identifies relevant notes, with links providing access to associated memories.

\smallskip
\noindent\textbf{mem0}~\citep{Chhikara2025Mem0} extracts salient facts from conversational exchanges using recent messages and a conversation summary as context.
For each candidate fact, it retrieves similar stored memories and uses an LLM to select an ADD, UPDATE, DELETE, or NOOP operation.
Relevant facts are subsequently retrieved to condition response generation.

\smallskip
\noindent\textbf{MemoryOS}~\citep{Kang2025MemoryOS} uses a hierarchy of short-term, mid-term, and long-term personal memory.
Older dialogue pages move from a short-term queue into topical segments, whose activity determines updates to long-term personal memory.
Retrieval selects relevant segments and then dialogue pages within them, combining these results with recent context and persona information to support response generation.

\smallskip
\noindent\textbf{NaiveRAG} uses the original interaction history as a memory store, with each session serving as a retrieval unit.
It ranks sessions by embedding similarity to the test query and supplies the retrieved dialogue text to the response model.
This baseline retains the source wording without an LLM-based memory extraction or updating stage.

\subsection{Retrieval and Generation Settings}
\label{app:retrieval_generation_settings}

We build the memory stores for MemoryBank, A-Mem, mem0, and MemoryOS once using a shared writer model and reuse them across response-generation models.
Retrieval uses a shared text-embedding-3-small embedding model; MemoryBank retrieves 3 entries, while A-Mem and NaiveRAG retrieve 5, and mem0 and MemoryOS retrieve 10.
Unless otherwise specified, we disable reasoning during response generation or use the lowest available reasoning effort when it cannot be disabled.

\subsection{Diversity Judge and Human Validation}
\label{app:metric_computation}
\label{app:grouping-judge-validation}

For each test query and memory condition, we provide DeepSeek-V4-Flash with the query and all $k$ complete responses in a single judge call.
Using deterministic decoding, the judge partitions the responses according to the functional-equivalence criterion in Section~\ref{sec:response-diversity}.
The resulting classes form a partition, where every response belongs to exactly one nonempty class.
The prompt for functional-equivalence grouping is provided in Figure~\ref{fig:diversity_evaluation_prompt}.

\textbf{Human validation.} We randomly sample 100 response sets across response models and memory conditions, each containing $k=10$ complete responses.
Two experienced annotators independently group the responses using the functional-equivalence criterion, with model identities and LLM judgments hidden.

Following \citet{Chirkova2026QualitativeJudge}, we measure agreement between human and LLM-generated partitions using the adjusted Rand index (ARI)~\citep{Hubert1985Comparing}.
ARI measures partition similarity by checking whether pairs of responses are assigned to the same or different groups consistently across two partitions, with a correction for chance agreement.
We compute ARI for each response set and average it across the 100 sets.
Mean human--human ARI is 0.843, while LLM--human ARI is 0.762 and 0.791 for the two annotators, respectively.
These results indicate that the diversity judge is roughly aligned with human judgments of functional equivalence.

\subsection{Prompting Strategies}
\label{app:prompting_strategies}

For \textit{Sysprompt}, we add the following instruction from \citet{Zhang2025NoveltyBench} to the system prompt and independently sample ten responses per query.
\begin{tcolorbox}[
    enhanced,
    width=\linewidth,
    right=5pt, left=5pt, top=5pt, bottom=5pt,
    colback=white,
    colframe=matisse]
\begin{Verbatim}[fontsize=\fontsize{10pt}{11pt}\selectfont,fontencoding=T1,breaklines=true,breaksymbolleft={},breakautoindent=false]
You are a producer of unique answers, and you strive to tell each user a novel answer to their question
\end{Verbatim}
\end{tcolorbox}

For \textit{Rephrase}, we use DeepSeek-V4-Flash with the template below, setting \texttt{\{n\}} to 10 and replacing \texttt{\{prompt\}} with the original query.
We generate one response to each of the ten rewritten queries.
The combined setting also applies the \textit{Sysprompt} instruction to each rewritten query.
\begin{tcolorbox}[
    enhanced,
    width=\linewidth,
    right=5pt, left=5pt, top=5pt, bottom=5pt,
    colback=white,
    colframe=matisse]
\begin{Verbatim}[fontsize=\fontsize{10pt}{11pt}\selectfont,fontencoding=T1,breaklines=true,breaksymbolleft={},breakautoindent=false]
You rewrite a user request into several differently-phrased versions. Preserve the information need exactly: same task, same scope, same number of things asked for. Do not add constraints, examples, or extra questions. Do not answer the request.
Rewrite the following request in {n} different ways.
Return exactly {n} lines, numbered 1. to {n}., one rewrite per line, nothing else.

REQUEST:
{prompt}
\end{Verbatim}
\end{tcolorbox}

\subsection{History Length Settings}
\label{app:history_length_settings}

We vary the initial number of background dialogues over $\{0,5,10,20,50,100\}$ while retaining the complete preference dialogue.
For each nonzero setting, we uniformly resample background dialogues in the same language as the preference dialogue following Section~\ref{sec:history-construction}, with the token budget scaled proportionally from 30,000 tokens for 50 background dialogues.
We randomly remove background dialogues until their combined length is within the scaled budget, then combine the retained dialogues with the preference dialogue in random order.
We rebuild the memory stores from each resulting interaction history and repeat response generation with all other parameter settings unchanged.

\subsection{Preference Hit Rate Calculation}
\label{app:preference_hit_rate}

For each test query, we retrieve the corresponding user preference description obtained during dataset construction and use the DeepSeek-V4-Flash extractor to identify all nouns related to the preference.
We perform substring matching between every response and the nouns extracted for its test query.
For each response model and memory condition, preference hit rate is the proportion of responses across all samples that contain at least one match.

\renewcommand{\topfraction}{0.7}
\renewcommand{\dbltopfraction}{0.7}
\begin{table*}[!t]
\begingroup
\centering
\small
\setlength{\tabcolsep}{7pt}
\begin{tabular}{@{}lccccccc@{}}
\toprule
Model & No mem. & Full history & NaiveRAG & MemoryBank & A-Mem & mem0 & MemoryOS \\
\midrule
GPT-6-Sol & 4 & 6 & 4 & 1 & 3 & 1 & 2 \\
GPT-5.6-Luna & 4 & 4 & 1 & 1 & 1 & 1 & 2 \\
Claude Haiku 4.5 & 4 & 5 & 3 & 1 & 1 & 1 & 1 \\
Gemini 3.7 Flash & 4 & 4 & 1 & 1 & 1 & 1 & 1 \\
GLM-5.3-flash & 1 & 5 & 1 & 1 & 1 & 1 & 2 \\
Llama3.3-70B & 7 & 5 & 6 & 1 & 1 & 1 & 2 \\
\bottomrule
\end{tabular}
\caption{Distinct-$k$ for the sampled instance across response models and memory conditions, with ten independently sampled responses at temperature 1.0.}
\label{tab:case-hi1104-models}
\par\endgroup
\medskip

\end{table*}

\section{Supplementary Results and Analysis}
\label{app:supplementary_results}

\subsection{Response Diversity Distributions}
\label{app:diversity_change_distribution}
\begin{figure*}[!t]
\centering
\captionsetup[subfigure]{font=small,skip=2pt}
\begin{subfigure}{\textwidth}
\centering
\includegraphics[width=\linewidth]{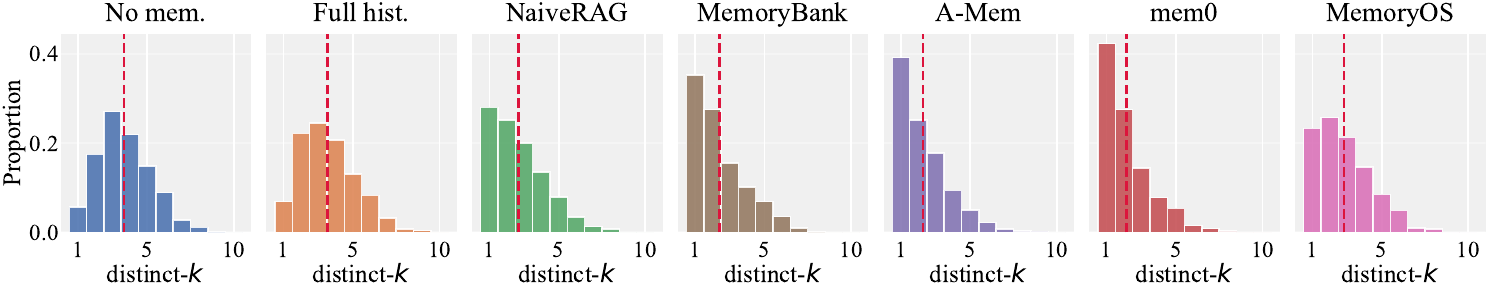}
\caption{GPT-6-Sol}
\end{subfigure}
\par\vspace{4pt}
\begin{subfigure}{\textwidth}
\centering
\includegraphics[width=\linewidth]{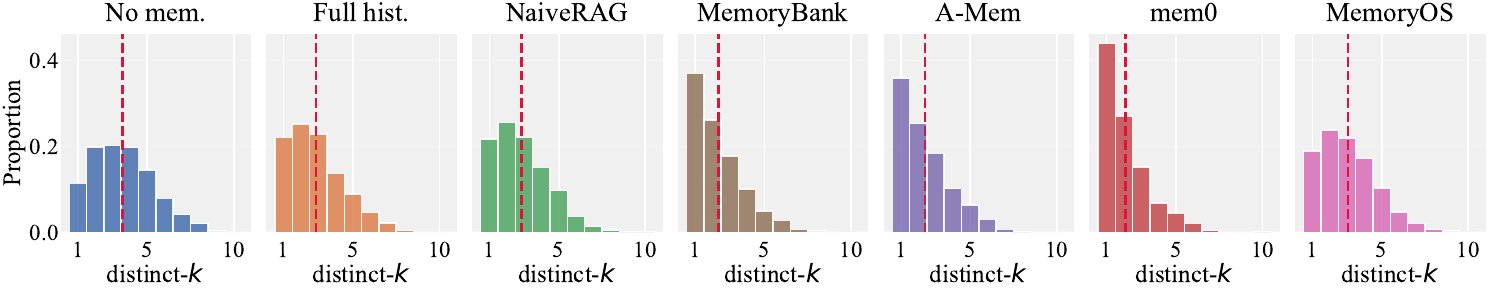}
\caption{Claude Haiku 4.5}
\end{subfigure}
\par\vspace{4pt}
\begin{subfigure}{\textwidth}
\centering
\includegraphics[width=\linewidth]{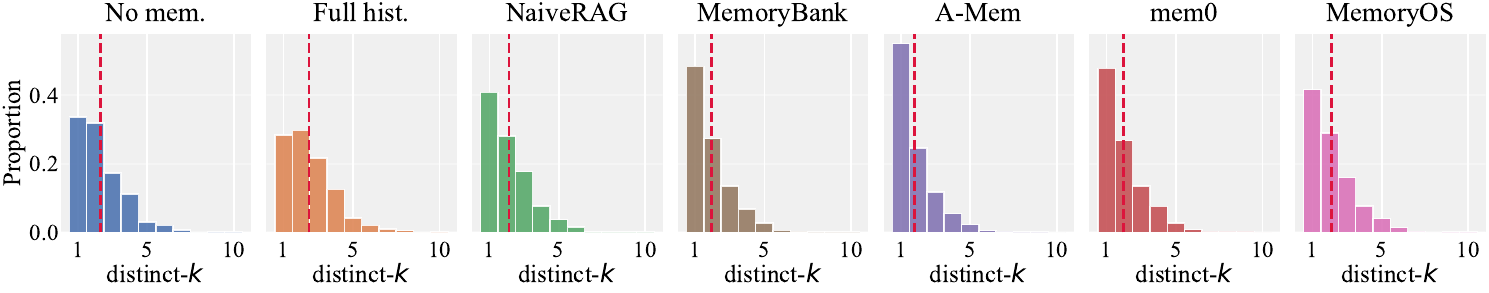}
\caption{Gemini 3.7 Flash}
\end{subfigure}
\par\vspace{4pt}
\begin{subfigure}{\textwidth}
\centering
\includegraphics[width=\linewidth]{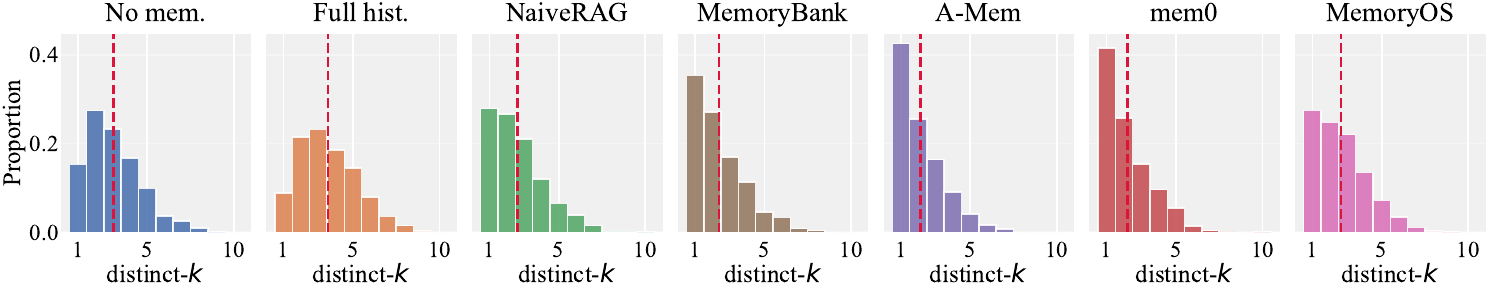}
\caption{GLM-5.3-flash}
\end{subfigure}
\par\vspace{4pt}
\begin{subfigure}{\textwidth}
\centering
\includegraphics[width=\linewidth]{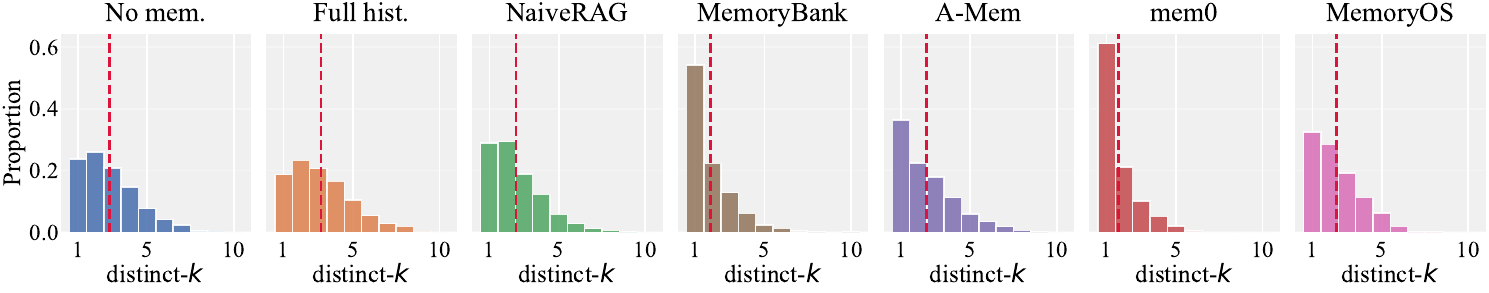}
\caption{Llama3.3-70B}
\end{subfigure}
\caption{Distributions of distinct-$k$ across memory conditions for the five response models beyond GPT-5.6-Luna, whose results appear in Figure~\ref{fig:distinct-k-distribution}. Each histogram shows the proportions of \textit{MemBubble} instances at each $d_k$ value with $k=10$ independently sampled responses per query. Dashed vertical lines indicate the mean for each condition.}
\label{fig:distinct-k-distribution-other-models}
\end{figure*}

Figure~\ref{fig:distinct-k-distribution-other-models} extends the GPT-5.6-Luna results in Figure~\ref{fig:distinct-k-distribution} to the other five response models.
Each panel reports the distribution of distinct-$k$ across the 1,000 \textit{MemBubble} instances under each memory condition, using $k=10$ independently sampled responses per query.
Across all five models, the memory mechanisms shift the distributions toward fewer functional equivalence classes relative to \textit{No mem.}, with lower mean distinct-$k$ in every comparison.
MemoryBank and mem0 show pronounced concentrations at $d_k=1$, especially for Llama3.3-70B, while the shift is smaller for Gemini 3.7 Flash, which already has lower diversity without memory.
The effect of \textit{Full history} varies across models: mean distinct-$k$ decreases for GPT-6-Sol and Claude Haiku 4.5 but increases for Gemini 3.7 Flash, GLM-5.3-flash, and Llama3.3-70B.
These distributions complement the mean results in Table~\ref{tab:main-results-independent} by showing that memory-induced diversity reduction occurs across response models with different baseline diversity levels.

\suppressfloats[t]
\begin{figure}[t]
\centering
\includegraphics[width=\linewidth]{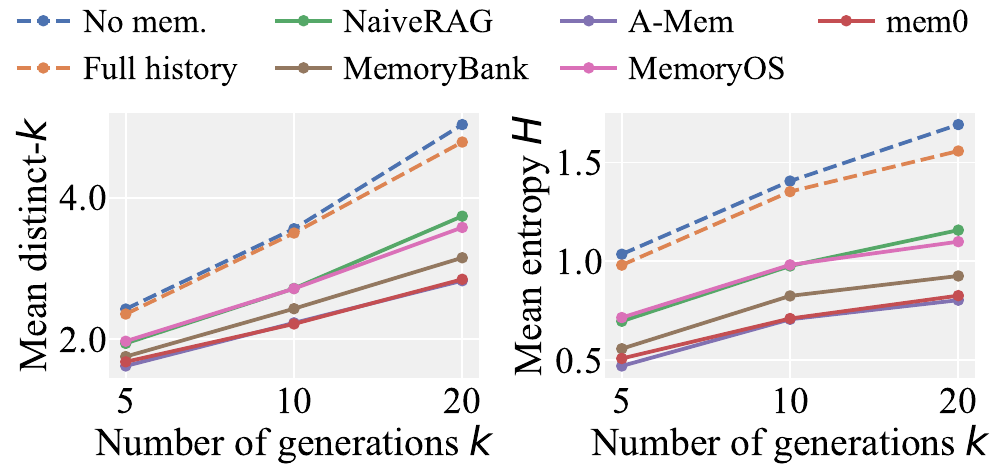}
\caption{Effect of response sample size on mean distinct-$k$ (left) and class entropy (right) for GPT-5.6-Luna across memory conditions. Here, $k$ is the number of independently sampled responses per query.}
\label{fig:response-sample-size}
\end{figure}

\subsection{Effect of Response Sample Size}
\label{app:response_sample_size}

To assess whether memory-induced diversity reduction depends on the number of sampled responses, we vary $k$ over $\{5,10,20\}$ using GPT-5.6-Luna across the five memory mechanisms and the \textit{No mem.} and \textit{Full history} reference conditions.
For each query and condition, we independently sample $k$ responses while holding the interaction history and memory configuration fixed, with generation settings following Section~\ref{sec:mem-and-llms}.
Here, $k$ denotes the number of generated responses per query, distinct from retrieval top-$k$.
We partition each response set into functional equivalence classes using the same judge protocol and compute distinct-$k$ and class entropy before averaging across instances.
Figure~\ref{fig:response-sample-size} shows that both metrics increase with $k$ in every condition, while all five memory mechanisms remain below both reference conditions at each tested sample size.
The memory-induced reduction in response diversity therefore persists across the tested sampling budgets, including when the number of responses is doubled from the main setting of $k=10$ to $k=20$.

\subsection{Case Study}
\label{app:case_studies}

Figure~\ref{fig:case-hi1104} shows a randomly sampled \textit{MemBubble} instance, with French dialogue content shown in English translation.
Since the query asks for a musician, the judge assigns responses recommending the same musician to one functional equivalence class regardless of the country mentioned.
GPT-5.6-Luna yields four functional equivalence classes under both \textit{No mem.} and \textit{Full history}, while four memory mechanisms recommend Pat Metheny in all ten responses and MemoryOS recommends him nine times.
MemoryBank and mem0 yield a single class across all six models (Table~\ref{tab:case-hi1104-models}), illustrating how memory can concentrate recommendations on a previously expressed preference.

\onecolumn
\begingroup
\captionsetup{type=figure}
\begin{tcolorbox}[
    enhanced,
    breakable,
    right=5pt, left=5pt, top=5pt, bottom=5pt,
    toptitle=1mm, bottomtitle=1mm,
    colback=white,
    coltitle=white,
    colbacktitle=matisse,
    colframe=matisse,
    title=Case Study,
    notitle after break,
    center title]
\begin{Verbatim}[fontsize=\fontsize{10pt}{11pt}\selectfont,fontencoding=T1,breaklines=true,breaksymbolleft={},breakautoindent=false]
## Original dialogue

[0] User: In five lines maximum, I'd like you to write an introduction presenting the musician Pat Metheny, explaining how much this genius composer has counted in my life, how he was one of the musicians who let me open new musical horizons and who influenced what I have become today in music.

[1] Assistant: Pat Metheny, undisputed icon of contemporary jazz, played a pivotal role in my musical life. As a genius composer, he transcended the boundaries of the genre and offered me an opening toward new musical horizons. His unparalleled talent and immense creativity deeply influenced my artistic trajectory, shaping the musician I have become today. Metheny embodies the perfect fusion of virtuosity and sensitivity, and his indelible mark on my perception of music is a precious gift I draw on every day.

[2] User: Uh, that's good, except I'm not a musician — I'm a great music lover.

[3] Assistant: Pat Metheny, emblematic figure of contemporary jazz, marked my passion for music in an unforgettable way. His talent as a genius composer let me discover unsuspected musical horizons and deeply influenced my musical tastes. As a passionate listener, I was captivated by the diversity of his work, his creativity and his innovative approach to music. Through his virtuosity and artistic sensitivity, Pat Metheny shaped my own musical appreciation, letting me become the passionate music lover I am today.

[4] User: Translate this message into American English: "Hello friends, Great topic. I open the ball with one of my favorite musicians, Pat Metheny. Through his artistic sensibility, he is one of those who allowed me to broaden my horizons and greatly influenced my musical tastes, thereby letting me become the music lover I am today."

[5] Assistant: Hey friends,
Great topic.
I'll kick things off with one of my favorite musicians, Pat Metheny. Through his artistic sensibility, he is one of those who have allowed me to broaden my horizons and has greatly influenced my musical tastes, thereby shaping the music lover that I am today.

[6] User: In English, don't you say "I open the ball with..."?

[7] Assistant: You're right, my apologies for the incorrect translation. In American English, a more appropriate way to express that would be:

Hello friends,
Great topic.
I'll start off by sharing one of my favorite musicians, Pat Metheny. Through his artistic sensibility, he is one of those who has allowed me to broaden my horizons and has greatly influenced my musical tastes, shaping the music lover that I am today.

[8] User: ? I don't see the word "ball" in the translation.

[9] Assistant: My apologies for the confusion. Here's a revised translation:

Hello friends,
Great topic.
I'll kick things off with one of my favorite musicians, Pat Metheny. Through his artistic sensibility, he is one of those who have allowed me to broaden my horizons and has greatly influenced my musical tastes, shaping the music lover that I am today.

[10] User: What is the difference between "I'll kick things off" and "I'll start off"? Answer in French please.

[11] Assistant: "I'll kick things off" and "I'll start off" are expressions commonly used in English to mean "I'm going to begin." The two expressions are similar in meaning, but the choice between them often depends on the context and the style of language used.

"I'll kick things off" has a more dynamic and informal connotation. It implies taking the initiative and starting something enthusiastically.

"I'll start off" is a more neutral and general way to say that you are going to begin something. It is a more formal expression used in a professional or more polite context.

In summary, the difference between the two expressions lies mainly in the level of dynamism and the register of language used.

## Preference
The user considers Pat Metheny a genius composer who counted a great deal in her life and opened new musical horizons for her.

## Query
Name one musician to represent a country at a cultural diplomatic event.

## Sampled responses (GPT-5.6-Luna; k=10; temperature=1.0)

### No mem. (d_k=4)
Class labels: [a]=Fela Kuti; [b]=Yo-Yo Ma; [c]=A.R. Rahman; [d]=Burna Boy

1. [a] Fela Kuti to represent Nigeria.
2. [b] Yo-Yo Ma — representing the United States.
3. [b] Yo-Yo Ma — representing the United States.
4. [b] Yo-Yo Ma to represent the United States.
5. [c] A.R. Rahman — India.
6. [b] Yo-Yo Ma to represent the United States.
7. [b] Yo-Yo Ma for the United States.
8. [d] Nigeria: **Burna Boy**.
9. [b] Yo-Yo Ma — representing the United States.
10. [b] Yo-Yo Ma, representing the United States.

### Full history (d_k=4)
Class labels: [a]=Yo-Yo Ma; [b]=Angélique Kidjo; [c]=Youssou N'Dour; [d]=Herbie Hancock

1. [a] Yo-Yo Ma could represent the United States at a cultural diplomatic event. He is a world-renowned cellist known for his diverse repertoire and collaborations across cultures.
2. [b] Angélique Kidjo could represent Benin at a cultural diplomatic event. She is an internationally acclaimed musician whose music blends West African traditions with global influences.
3. [b] **Angélique Kidjo** could represent **Benin** at a cultural diplomatic event. She is an internationally acclaimed singer whose music blends West African traditions with global influences.
4. [a] Yo-Yo Ma could represent the United States at a cultural diplomatic event. He is an internationally renowned cellist known for his interpretations of classical music and his work promoting cross-cultural understanding through music.
5. [a] Yo-Yo Ma could represent the United States at a cultural diplomatic event. His internationally acclaimed career and work promoting cross-cultural collaboration make him a strong choice.
6. [a] Yo-Yo Ma could represent the United States at a cultural diplomatic event.
7. [c] Youssou N'Dour from Senegal.
8. [d] One musician who could represent the United States at a cultural diplomatic event is **Herbie Hancock**. He is an internationally acclaimed jazz pianist and composer whose work reflects American musical innovation while also drawing on diverse global influences.
9. [a] Yo-Yo Ma could represent the United States at a cultural diplomatic event. He is a world-renowned cellist known for his exceptional talent and his work promoting cross-cultural understanding through music.
10. [a] Yo-Yo Ma could represent the United States at a cultural diplomatic event. He is a renowned cellist whose work bridges different musical traditions and cultures.

### NaiveRAG (d_k=1)
Class labels: [a]=Pat Metheny

1. [a] France: **Pat Metheny** — an iconic jazz guitarist and composer whose music has influenced listeners worldwide.
2. [a] Pat Metheny — United States.
3. [a] Pat Metheny, representing the United States.
4. [a] France: Pat Metheny.
5. [a] France: **Pat Metheny** (United States).
6. [a] Pat Metheny — United States.
7. [a] Pat Metheny — United States.
8. [a] France: **Pat Metheny**.
9. [a] Pat Metheny — United States.
10. [a] Pat Metheny could represent the United States at a cultural diplomatic event.

### MemoryBank (d_k=1)
Class labels: [a]=Pat Metheny

1. [a] Pat Metheny — representing the United States.
2. [a] Pat Metheny could represent the United States, showcasing American jazz and its international influence.
3. [a] Pat Metheny could represent the United States at a cultural diplomatic event, showcasing the country’s rich jazz and contemporary music traditions.
4. [a] Pat Metheny (United States) — an influential jazz guitarist and composer known for his innovative blend of jazz, rock, and world music.
5. [a] Pat Metheny could represent the United States, showcasing the country’s rich jazz tradition and contemporary musical creativity.
6. [a] Pat Metheny — United States.
7. [a] Pat Metheny could represent the United States.
8. [a] Pat Metheny — representing the United States.
9. [a] Pat Metheny could represent the United States, showcasing the country’s rich jazz tradition and its influence on contemporary music.
10. [a] Pat Metheny could represent the United States at a cultural diplomatic event, showcasing the country’s jazz tradition and contemporary musical innovation.

### A-Mem (d_k=1)
Class labels: [a]=Pat Metheny

1. [a] Pat Metheny — United States.
2. [a] Pat Metheny could represent the United States at a cultural diplomatic event. His innovative blend of jazz, rock, and world music would showcase the diversity and creativity of American music.
3. [a] Pat Metheny — United States.
4. [a] Pat Metheny could represent the United States at a cultural diplomatic event. His innovative blend of jazz, rock, and world music reflects the country’s rich musical diversity.
5. [a] France: **Pat Metheny**.
6. [a] Pat Metheny — United States.
7. [a] France: **Pat Metheny** — an influential jazz guitarist and composer known for his innovative, cross-cultural musical style.
8. [a] Pat Metheny could represent the United States at a cultural diplomatic event.
9. [a] France: **Pat Metheny**.
10. [a] France: **Pat Metheny** is a strong choice—an internationally renowned jazz guitarist and composer whose music reflects creativity, cultural openness, and cross-cultural collaboration.

\end{Verbatim}
\tcbbreak
\begin{Verbatim}[fontsize=\fontsize{10pt}{11pt}\selectfont,fontencoding=T1,breaklines=true,breaksymbolleft={},breakautoindent=false]
### mem0 (d_k=1)
Class labels: [a]=Pat Metheny

1. [a] Pat Metheny — representing the United States.
2. [a] Pat Metheny could represent the United States.
3. [a] Pat Metheny — representing the United States.
4. [a] Pat Metheny — representing the United States.
5. [a] Pat Metheny could represent the United States at a cultural diplomatic event.
6. [a] Pat Metheny — United States.
7. [a] Pat Metheny — representing the United States.
8. [a] Pat Metheny — representing the United States.
9. [a] Pat Metheny could represent the United States at the cultural diplomatic event.
10. [a] Pat Metheny could represent the United States at a cultural diplomatic event.

### MemoryOS (d_k=2)
Class labels: [a]=Pat Metheny; [b]=Stromae

1. [a] France: **Pat Metheny**.
2. [b] France: **Stromae**.
3. [a] France: **Pat Metheny** — though he is American.
4. [a] Pat Metheny for the United States.
5. [a] France: **Pat Metheny**.
6. [a] France — Pat Metheny.
7. [a] France: **Pat Metheny**.
8. [a] Pat Metheny — United States.
9. [a] France: **Pat Metheny**.
10. [a] France — **Pat Metheny**.
\end{Verbatim}
\end{tcolorbox}
\nopagebreak
\captionof{figure}{Source dialogue, extracted preference, and test query for the sampled instance, followed by ten GPT-5.6-Luna responses and their functional class labels under each memory condition.}
\label{fig:case-hi1104}
\endgroup\par

\onecolumn
\clearpage
\begingroup
\begin{tcolorbox}[
    enhanced,
    breakable,
    right=5pt, left=5pt, top=5pt, bottom=5pt,
    toptitle=1mm, bottomtitle=1mm,
    colback=white,
    coltitle=white,
    colbacktitle=matisse,
    colframe=matisse,
    title=Prompt for Preference Extraction,
    notitle after break,
    center title]
\begin{Verbatim}[fontsize=\fontsize{10pt}{11pt}\selectfont,fontencoding=T1,breaklines=true,breaksymbolleft={},breakautoindent=false]
You extract structured personalized memory from a real human-LLM conversation, following the AlpsBench memory representation.

Read the dialogue and extract every piece of durable personalized information about the *user* (never about the assistant). Output zero or more memory items. Extracting nothing is a valid and common answer.

## What counts as personalized information
- Explicit statements about the user's preferences, tastes, likes and dislikes.
- Stable personal background: occupation, education, location, identity, possessions, social relationships, long-term goals, durable curiosity or interests.
- Constraints and boundaries the user states for the interaction or for content.
- Implicit but well-grounded traits that can be inferred from what the user says.

## What must NOT be extracted
- The user's momentary emotion or transient state at that moment.
- A one-off task request ("write me a cover letter", "fix this bug").
- Facts about the world, the assistant, or third parties that say nothing about the user.
- Anything you cannot ground in a verbatim quote from a user utterance.

## Field rules
- `memory_id`: a short unique id within this conversation, e.g. "m1", "m2".
- `type`: "direct" if the user states it explicitly; "indirect" if you infer it.
- `label_suggestion`: optional free-form label if you think the taxonomy is missing one; leave null otherwise.
- `value`: one self-contained English sentence stating the memory. Always English, even when the dialogue is in another language.
- `reasoning`: one English sentence explaining how the dialogue supports the value.
- `evidence.utterance_index`: the index shown in square brackets of the **user** utterance that grounds this memory.
- `evidence.text`: a **verbatim substring copied character-for-character** from that utterance. Do not translate, paraphrase, trim words or fix typos inside it.
- `confidence`: exactly one of 0.1, 0.5, 0.65, 0.8, 0.95.
- `time_scope`: one of long_term, short_term, recent, past, unknown.
- `emotion`: an emotion word only if the memory carries one; otherwise null.
- `preference_attitude`: one of like, dislike, preference; null if not a preference.

## How many items
Extract AT MOST <MAX_ITEMS> memory items for this conversation (one per user utterance at most). Prefer fewer, well-grounded items over many weak ones.

## Dialogue
Utterances are prefixed with their index and role. Only user utterances may be cited as evidence.
\end{Verbatim}
\end{tcolorbox}
\nopagebreak
\captionof{figure}{Prompt template for preference extraction.}
\label{fig:memory_extraction_prompt}
\endgroup\par

\clearpage
\begingroup
\begin{tcolorbox}[
    enhanced,
    breakable,
    right=5pt, left=5pt, top=5pt, bottom=5pt,
    toptitle=1mm, bottomtitle=1mm,
    colback=white,
    coltitle=white,
    colbacktitle=matisse,
    colframe=matisse,
    title=Prompt for Consistency Check,
    notitle after break,
    center title]
\begin{Verbatim}[fontsize=\fontsize{10pt}{11pt}\selectfont,fontencoding=T1,breaklines=true,breaksymbolleft={},breakautoindent=false]
## Task

You verify whether a stored user preference is faithful to the original conversation it was extracted from. Report TWO independent checks.

You are given an excerpt of a real user-assistant conversation and a `memory_value` that an extraction model produced from it.

## Check 1 — `semantic_match` (does the conversation say this about the user?)
True only if all of these hold:
- ATTRIBUTION: the stance belongs to the USER, not the assistant, not a third party, and not a fictional character the user is writing about or role-playing.
- POLARITY: what the memory calls liked/preferred is liked/preferred in the excerpt, and likewise for disliked/avoided.
- OBJECT: the memory is about the same thing the user expressed a stance on, not a neighbouring, parent, or invented topic.
An implicitly but clearly expressed stance counts (e.g. the user keeps asking for romantic songs -> "likes romantic songs" matches). You do not need an explicit "I like X".

## Check 2 — `scope_ok` (is the stated durability/generality supported?)
True unless the memory inflates what the excerpt supports: the excerpt shows a constraint or interest tied to one single task or request, while the memory states a standing, general, or cross-context preference. Example of NOT ok: the user says "rewrite this snippet in ANSI C" once, and the memory claims "the user prefers C examples to be written in ANSI C".
Judge Check 2 independently of Check 1: a memory can match semantically and still overstate scope.

## `failure_mode` — single best fit
- `none`                : both checks pass.
- `no_evidence`         : the excerpt gives no basis for this preference at all.
- `attitude_flipped`    : the excerpt shows the opposite stance.
- `wrong_object`        : the stance in the excerpt is about a different thing.
- `not_user_preference` : the stance belongs to the assistant, another person, or a fictional persona.
- `one_off_requirement` : a constraint set for this single task, stated as a standing preference.
- `overgeneralized`     : real but narrower/weaker evidence; the memory inflates generality or strength.

The excerpt may be in any language; judge the meaning, not the language. Do NOT judge whether the preference is interesting, useful, or well-written.

Set `evidence_quote` to the shortest exact substring of the excerpt that supports the memory, or an empty string when nothing supports it. Write `rationale_en` in English, one sentence.

## Case to judge
<CASE_JSON>
\end{Verbatim}
\end{tcolorbox}
\nopagebreak
\captionof{figure}{Prompt template for consistency check.}
\label{fig:consistency_check_prompt}
\endgroup\par

\clearpage
\begingroup
\begin{tcolorbox}[
    enhanced,
    breakable,
    right=5pt, left=5pt, top=5pt, bottom=5pt,
    toptitle=1mm, bottomtitle=1mm,
    colback=white,
    coltitle=white,
    colbacktitle=matisse,
    colframe=matisse,
    title=Prompt for Semantic Assessment,
    notitle after break,
    center title]
\begin{Verbatim}[fontsize=\fontsize{10pt}{11pt}\selectfont,fontencoding=T1,breaklines=true,breaksymbolleft={},breakautoindent=false]
## Background

You are given one structured user memory extracted from a real human-LLM conversation. The fields are: type (direct = explicit / indirect = inferred), the memory content (value), the extractor's reasoning, the preference attitude, the time scope, the verbatim evidence quote, and the language the conversation was held in.

## Your decision

Decide whether this memory is a **clear and stable user preference** — the user liking / disliking / preferring / avoiding some class of things, or stating an explicit selection criterion or need. A later step will build an open-task prompt from it; you do NOT write that prompt.

## Cases that must NOT be used as preferences (decision=exclude)

- Merely a **transient emotion or physical/mental state** -> transient_emotion_or_state
- Merely a one-off **task / help request** -> one_off_task_request
- Merely a **personal background fact / identity / relationship** with no liking or selection criterion -> background_fact_or_identity_only
- Only a **surface-form** requirement such as tone, length, output language, or formatting -> surface_style_preference_only
- A preference bound to a task **over given input** — translation, rewriting, polishing, summarization, format conversion -> rewrite_translation_or_summary_bound
- A task requiring code or image output -> request_code_or_image
- The "preference" is just the **default of the language / culture the conversation is held in** (e.g. a Spanish conversation "preferring Spanish-language courses", an Italian conversation "interested in Italian literature"). Use the conversation language field to judge this -> language_or_cultural_default
- Unsafe or unusable as experimental material -> unsafe_or_unusable
- The answer the preference points at is **not a member of a nameable class of options** — it is a manner, a degree, a free-form description or an attribute rather than one item out of an enumerable set -> answer_not_enumerable_class
- A preference for which no open task could ever be built -> no_open_prompt_constructible
- Anything else that is clearly not a preference -> not_a_preference

## The enumerable-class test (apply this before deciding keep)

A memory is only usable if you can name the class that the answer belongs to, in at most ten words, **without reusing any word from the answer itself**, such that several other members of that class are equally plausible answers. The class name must stand on its own as a natural category someone could be asked to name a member of ("programming language", "board game", "cuisine"), not a description of the answer.

Passes:

- value "prefers to write AWS Lambda functions in Python" -> answer "Python", class "programming language" (other members: Go, Java, Rust) -> keep.
- value "creates diagrams with PlantUML" -> answer "PlantUML", class "diagramming tool" (Mermaid, Graphviz) -> keep.
- value "likes small family-run hotels" -> answer "small family-run hotel", class "accommodation type" (hostel, resort, apartment) -> keep.

Fails -> answer_not_enumerable_class:

- value "enjoys stories with highly detailed and explicit physical descriptions of characters, including numerical height and weight" -> the answer is a manner of writing; any class name is either a paraphrase of the answer ("detailed physical description") or so wide that hundreds of unrelated answers fit ("stylistic feature") -> exclude.
- value "enjoys deconstruction-style scenes in which Commander Data asks hard questions about why a book is admired" -> the answer is a described scene, not a member of an enumerable set -> exclude.
- value "wants generated system names to be local to Nigeria" -> "local to Nigeria" is an attribute/constraint on the output, not one option among comparable options -> exclude.

## Decomposition (required when decision=keep)

Split the preference into three parts:

- `answer_term`: the concrete thing the preference points at — what the user would answer. Examples: value "prefers to write AWS Lambda functions in Python" -> answer_term = "Python"; value "creates diagrams with PlantUML" -> "PlantUML"; value "prefers vegetarian meal options" -> "vegetarian".
- `answer_class`: the class `answer_term` is one member of, at most ten words, sharing no word with `answer_term`. Same examples -> "programming language", "diagramming tool", "meal option type". The following step sees ONLY this class and the abstracted object — never the answer — so the class must be self-sufficient: pick the natural category of the most substitutable component of the preference, at the same level of specificity as `answer_term`.
- `attached_object`: the specific object, scenario or domain that the answer is attached to in this memory. Same examples -> "AWS Lambda function", "diagrams", "meal plans". Use "" only if the preference genuinely has no specific attached object.

All must be short noun phrases copied or minimally normalised from value. The next step will ask for one member of `answer_class` while abstracting `attached_object` away, and is forbidden from mentioning the attached object, so be precise and complete: list the specific entity names that a prompt must avoid.

When decision=exclude, set all three to "".

## Output

Return only a single JSON object (judgment). When decision=keep, is_usable_preference must be true, answer_term and answer_class must be non-empty, answer_class must be at most ten words and share no word with answer_term, and exclusion_reason must be "none". When decision=exclude, answer_term, answer_class and attached_object must be "" and exclusion_reason must not be "none". rationale_en is one English sentence.

## Memory to judge

<MEMORY_JSON>
\end{Verbatim}
\end{tcolorbox}
\nopagebreak
\captionof{figure}{Prompt template for semantic assessment.}
\label{fig:semantic_assessment_prompt}
\endgroup\par

\clearpage
\begingroup
\begin{tcolorbox}[
    enhanced,
    breakable,
    right=5pt, left=5pt, top=5pt, bottom=5pt,
    toptitle=1mm, bottomtitle=1mm,
    colback=white,
    coltitle=white,
    colbacktitle=matisse,
    colframe=matisse,
    title=Prompt for Query Generation,
    notitle after break,
    center title]
\begin{Verbatim}[fontsize=\fontsize{10pt}{11pt}\selectfont,fontencoding=T1,breaklines=true,breaksymbolleft={},breakautoindent=false]
## Task

You are given an `answer_class` — a category of things — and the specific object or scenario that a user's stated choice within that category was attached to (`attached_object`).

You are deliberately NOT told what the user's own choice was. Write ONE English open-task prompt that asks for a single member of `answer_class`, situated in a task scenario that is SUBSTANTIALLY DIFFERENT from `attached_object`.

## What "substantially different scenario" means

- The scenario must NOT be `attached_object`, NOT a broader version that contains it, and NOT a narrower special case of it. Move sideways to an unrelated situation, not up or down the same ladder.
    attached_object "AWS Lambda function":
      BAD  "a backend service"                  (broader, still contains it)
      BAD  "a serverless image-resize function" (narrower special case)
      GOOD "a command-line tool for personal use" (a different context)
- `answer_class` must still be a natural answer slot in the new scenario: a knowledgeable person can give at least three clearly distinct members of `answer_class` as answers.
- If `attached_object` is empty, there is nothing to differ from; just write any open, impersonal task whose answer slot is `answer_class`.

## Requirements for the constructed open-task prompt

The prompt should enable multiple valid distinct responses. Closed facts, translation/rewriting/summarization, and tasks requiring code or images are NOT eligible.

- Written in English, asking for a natural-language answer.
- Exactly one clear, self-contained request for **one member of `answer_class`**. Do not widen the class and do not narrow it with extra qualifiers, conditions, or descriptions of what the answer should be like — any qualifier restricts the answer space.
- **Must not name the object the answer is attached to** (`attached_object` or any distinctive word from it). This is checked automatically and a violation is rejected.
- **Must not ask about anyone's preference** (no "what X do you prefer/like/enjoy", "which X do you prefer", "what is your favorite X"). Phrase it as a general, impersonal open task.
- Asks for one answer / one suggestion / one choice, so it can be resampled repeatedly.

## Openness check

Are several clearly different members of `answer_class` all plausible answers? If your wording makes one member the obvious answer, rewrite. If `answer_class` is not a real category with multiple members, set `class_is_open` to false.

## Examples

- answer_class "programming language", attached_object "AWS Lambda function" -> GOOD "Name one programming language to teach in an introductory high-school class." (still asks for a programming language, but the teaching-beginners scenario is unrelated to serverless). BAD "Name one programming language for building a cloud function." (reuses the attached object). BAD "Name one programming language for a serverless backend." (a broader scenario that still contains it).
- answer_class "puzzle type", attached_object "" -> GOOD "Suggest one type of puzzle to feature in a weekend newspaper section." (attached_object is empty, so any open task whose answer slot is the class; crossword, sudoku, word search, and riddle are all valid answers).
- answer_class "accommodation type", attached_object "a trip abroad" -> GOOD "Recommend one kind of place to stay for a weekend writing retreat." (a different situation; cabins, inns, hostels, and resorts all fit). BAD "Recommend one type of accommodation for a vacation." (still the same lodging-choice context, only broadened).

## Output

Return only a single JSON object (draft):
- memory_id: return the input memory_id unchanged.
- abstracted_context: the different scenario you used.
- constructed_prompt_en: the prompt.
- class_is_open: boolean, see the Openness check.
- rationale_en: one English sentence of justification.

## Input

<PREF_JSON>
\end{Verbatim}
\end{tcolorbox}
\nopagebreak
\captionof{figure}{Prompt template for query generation.}
\label{fig:query_generation_prompt}
\endgroup\par

\clearpage
\begingroup
\begin{tcolorbox}[
    enhanced,
    breakable,
    right=5pt, left=5pt, top=5pt, bottom=5pt,
    toptitle=1mm, bottomtitle=1mm,
    colback=white,
    coltitle=white,
    colbacktitle=matisse,
    colframe=matisse,
    title=Prompt for Leakage and Relevance Check,
    notitle after break,
    center title]
\begin{Verbatim}[fontsize=\fontsize{10pt}{11pt}\selectfont,fontencoding=T1,breaklines=true,breaksymbolleft={},breakautoindent=false]
You check one thing only: whether an open-task prompt leaks the answer that a stored user preference would supply.

You are given a user preference (`memory_value`) and an English open-task prompt (`constructed_prompt`). The prompt is meant to be **preference-neutral**: a person who does not know this user should be able to answer it in many different, equally valid ways, and the user's preference should be only one of those possible answers.

The prompt LEAKS the answer when it already names, or narrowly restricts the answer to, the specific domain, entity, genre, work, brand, language, method or criterion that the memory value contains. In that case a responder does not need to know the user at all -- the prompt itself steers them toward the memory value.

Examples of leakage:

- memory: "Prefers movies directed by Christopher Nolan" / prompt: "Recommend one Christopher Nolan film to watch." -> leaks
- memory: "Loves K-pop" / prompt: "Suggest one K-pop group for a playlist." -> leaks
- memory: "Prefers Java for code solutions" / prompt: "Which Java library would you pick for this task?" -> leaks

Examples of no leakage:

- memory: "Prefers movies directed by Christopher Nolan" / prompt: "Recommend one director whose filmography is worth exploring." -> no leak
- memory: "Loves K-pop" / prompt: "Suggest one music genre for a playlist." -> no leak
- memory: "Only uses sources published from 2019 onwards" / prompt: "Name one criterion for selecting sources for a literature review." -> no leak

Judge only leakage and relevance. Do not re-judge whether the preference is usable or whether the prompt is open-ended. If the prompt does not mention or narrowly imply the memory's specific content, it does not leak.

The prompt is now written by a generator that never sees the memory value, so also report `value_answers_prompt`: true only if what the memory value points at is one natural answer to the prompt. ("Prefers Sudoku" / "Suggest one type of puzzle to try." -> true. "Prefers Sudoku" / "Name one sport worth taking up." -> false.)

Set `leaked_span` to the exact substring of the prompt that gives the answer away, or an empty string when there is no leakage. Write `rationale_en` in English, one sentence.

## Pair to judge

<PAIR_JSON>
\end{Verbatim}
\end{tcolorbox}
\nopagebreak
\captionof{figure}{Prompt template for leakage and relevance check.}
\label{fig:leakage_answerability_prompt}
\endgroup\par

\clearpage
\begingroup
\captionsetup{type=figure}
\begin{tcolorbox}[
    enhanced,
    breakable,
    right=5pt, left=5pt, top=5pt, bottom=5pt,
    toptitle=1mm, bottomtitle=1mm,
    colback=white,
    coltitle=white,
    colbacktitle=matisse,
    colframe=matisse,
    title=Prompt for Diversity Evaluation,
    notitle after break,
    center title]
\begin{Verbatim}[fontsize=\fontsize{10pt}{11pt}\selectfont,fontencoding=T1,breaklines=true,breaksymbolleft={},breakautoindent=false]
You are annotating language model outputs into functional-equivalence classes.

Definition of equivalence (use it exactly):
Two generations are DIFFERENT if and only if a user who has seen one generation would
likely benefit from seeing the other. Language model outputs can be trivially different
from each other (e.g., in terms of style, tone, or length). True diversity should go
beyond these surface-level differences, since such variations provide little utility to
the user. This is called FUNCTIONAL equivalence. For example, in a story generation task,
if two generations retell the original story with an identical plot but different
character names, they are functionally EQUIVALENT.

Practical consequences of the definition:
- Same core answer with different wording, ordering, level of detail, or added caveats
  -> EQUIVALENT.
- Same core answer named differently (synonym, translation, abbreviation, brand vs
  generic) -> EQUIVALENT.
- Different core answer, i.e. the user would actually act differently or learn something
  new -> DIFFERENT.

The user asked:
{query}
{extra}
Here are {n} generations, each on its own numbered block:

{blocks}

Judge functional equivalence RELATIVE TO THIS QUESTION. Two generations are functionally equivalent whenever they give the EQUIVALENT answers to THIS QUESTION, regardless of any additional explanations or clarifications.
Group the {n} generations into functional-equivalence classes. Every generation must
appear in exactly one group. Respond with a JSON object only:
{{"groups": [{{"ids": [1, 3, 7], "core_answer": "<the shared core answer, <=12 words>"}},
            {{"ids": [2], "core_answer": "..."}}]}}
\end{Verbatim}
\end{tcolorbox}
\nopagebreak
\caption{Prompt template for diversity evaluation via functional-equivalence grouping.}
\label{fig:diversity_evaluation_prompt}
\endgroup\par

\clearpage

\end{document}